\documentclass{article} 
\usepackage{iclr2027_conference,times}

\usepackage{amsmath,amsfonts,bm}

\def\eqref#1{equation~\ref{#1}}

\def\1{\bm{1}}

\DeclareMathAlphabet{\mathsfit}{\encodingdefault}{\sfdefault}{m}{sl}
\SetMathAlphabet{\mathsfit}{bold}{\encodingdefault}{\sfdefault}{bx}{n}

\usepackage{url}

\usepackage{tabularx}
\usepackage{array}

\usepackage{graphicx}
\usepackage{booktabs}
\usepackage{makecell}
\usepackage[table]{xcolor}
\usepackage{fontawesome5}
\usepackage{multirow}
\definecolor{myfire}{HTML}{ea3323}
\definecolor{darkgreen}{RGB}{0, 154, 85}
\usepackage{pifont}
\usepackage{tikz}
\usepackage{gradient-text}

\usepackage{tabularx}
\usepackage{array}
\usepackage{multirow}
\usepackage{booktabs}
\usepackage[table]{xcolor}

\usepackage[titles]{tocloft}
\usepackage{wrapfig}
\usepackage{etoc}
\usepackage{setspace}
\newsavebox{\logobox}
\sbox{\logobox}{\includegraphics[width=0.07\textwidth]{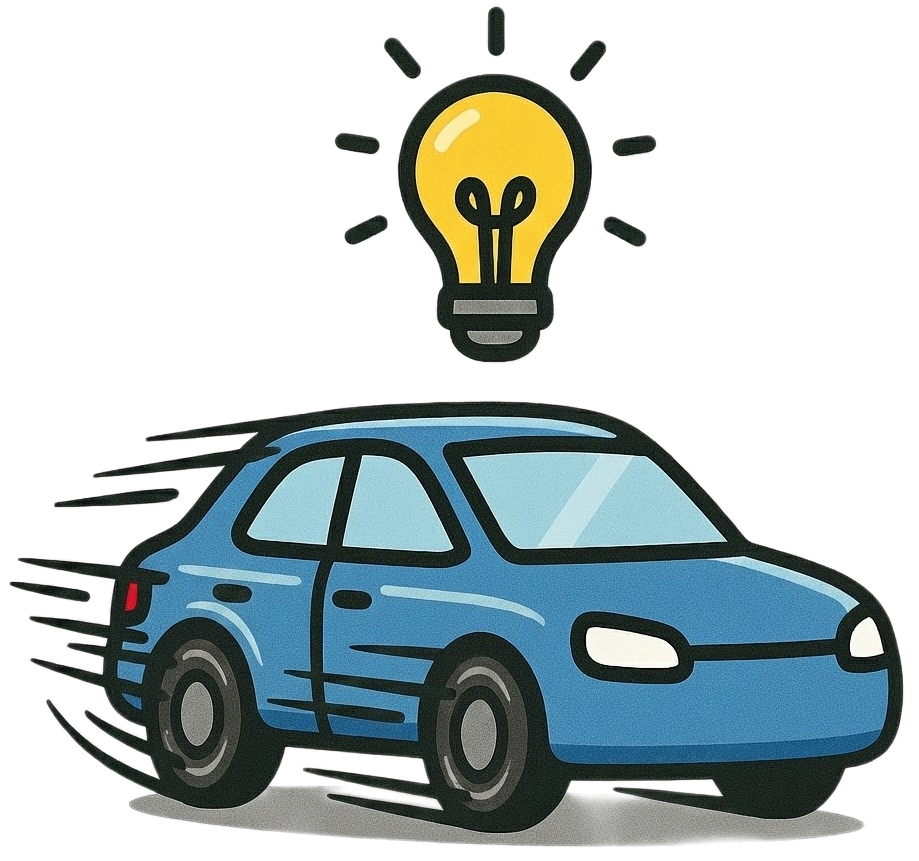}}

\usepackage[colorlinks=true,
            linkcolor=orange,
            citecolor=orange,
            urlcolor=orange]{hyperref}
\usepackage{cleveref}

\newcommand{\spanvla}{%
    \textsc{SpanVLA}%
}

\title{\raisebox{-.18\height}{\includegraphics[width=0.07\textwidth]{figure/logo.png}} \hspace{-1.1em} \textbf{\gradientRGB{SpanVLA}{9,29,83}{23,99,180}}: Learning from Negative-Recovery Samples with Fast Action Bridging for Vision-Language-Action Model}

\author{%
\hspace*{-\tabcolsep}%
\parbox{\textwidth}{%
\raggedright
\vspace{0.3cm}
\mbox{Zewei Zhou$^{1,2}$\textsuperscript{*}}%
\quad
\mbox{Ruining Yang$^{2,3}$\textsuperscript{*}}
\quad
\mbox{Tony (Xuewei) Qi$^{2}$\textsuperscript{†}}%
\quad
\mbox{Yiluan Guo$^{2}$}
\quad
\mbox{Sherry X. Chen$^{2}$}
\quad
\mbox{Tao Feng$^{2}$}
\quad
\mbox{Kateryna Pistunova$^{2}$}
\quad
\mbox{Yishan Shen$^{2}$}
\quad
\mbox{Lili Su$^{3}$}
\quad
\mbox{Jiaqi Ma$^{1}$}
\\[4pt]
\normalfont
$^{1}$UCLA
\quad
$^{2}$Motional
\quad
$^{3}$NEU
\quad\quad\quad
\textsuperscript{*} Equal contribution
\quad
\textsuperscript{†} Corresponding author
}%
\hspace*{-\tabcolsep}%
}

\iclrfinalcopy 
\begin{document}

\maketitle

\fancyhead{}
\fancyhead[L]{\raisebox{-.12\height}{\includegraphics[width=0.04\textwidth]{figure/logo.png}} \hspace{0.1em} \textsc{SpanVLA}}
\renewcommand{\headrulewidth}{0.4pt}

\begin{figure}[h]
    \centering
    \vspace{-0.7cm}
    \includegraphics[width=0.99\linewidth]{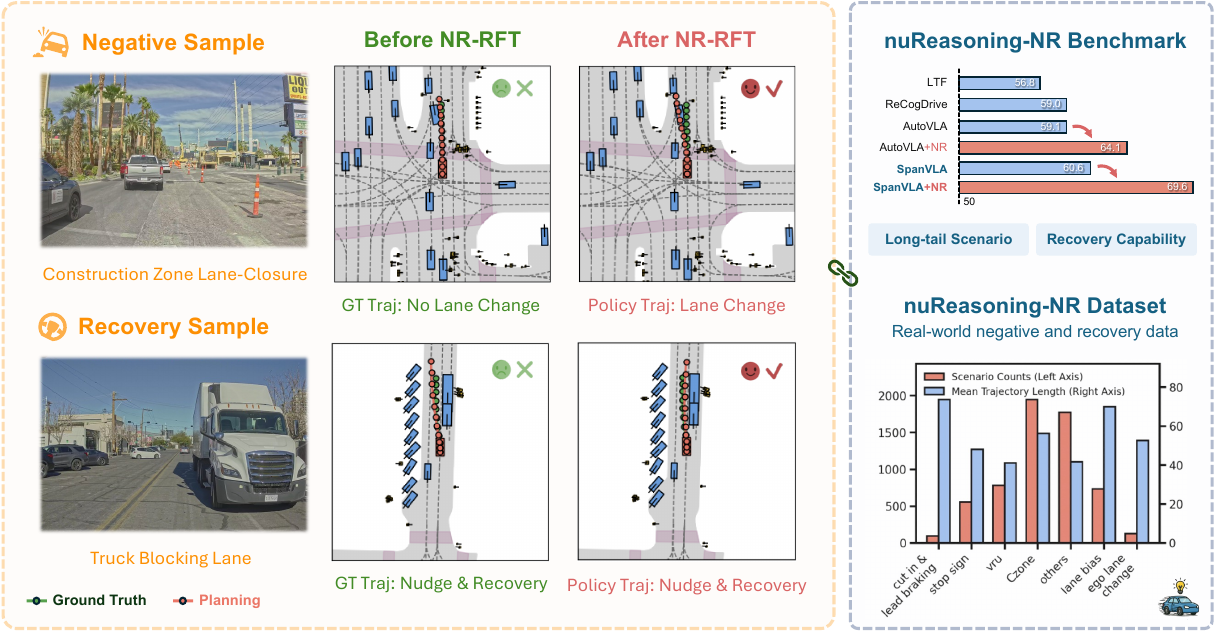}
    \vspace{-0.05cm}
    \caption{\textbf{Learning what to avoid and how to recover.} We introduce nuReasoning-NR, the first real-world driving dataset and benchmark for learning from negative and recovery samples. With the negative and recovery supervision, \spanvla{} predicts an earlier merge before a lane closure and a return to the original lane after passing an obstruction. \textbf{Website:} \footnotesize \url{https://spanvla.github.io/}}
    \label{fig:data_distribution}
\end{figure}

\begin{abstract}
    \vspace{-0.25cm}
    Vision-Language-Action (VLA) models bring rich world knowledge and reasoning capabilities to autonomous driving. However, most driving VLAs are trained only on positive expert demonstrations, which teach models what to do but provide limited supervision on which behaviors to avoid and how to recover from these failures, leading to limited robustness. We introduce \textsc{\textbf{\gradientRGB{SpanVLA}{9,29,83}{23,99,180}}}, a robust and efficient driving VLA, together with \textbf{nuReasoning-NR}, the first real-world driving dataset and benchmark for learning from negative and recovery samples. An extension of nuReasoning, it comprises 36K reasoning-intensive driving scenarios, including 3K suboptimal trajectories and 3K expert recovery trajectories. To exploit their asymmetric supervision, we introduce \textbf{Negative--Recovery Reinforcement Fine-Tuning (NR-RFT)}, which penalizes undesirable behaviors and rewards expert corrective actions. Because challenging scenarios produce rollout groups dominated by similar suboptimal or even infeasible trajectories, it further incorporates an advantage correction to preserve informative optimization signals and encourage exploration. For efficient action generation, \spanvla{} retains autoregressive reasoning and introduces a lightweight flow-matching action bridge conditioned on sparse-layer KV caches and initialized from the historical trajectory. Extensive experiments on NAVSIM v1/v2, nuReasoning, and the nuReasoning-NR benchmark demonstrate strong and robust planning performance and efficient action generation. Qualitative results across diverse challenging scenarios further highlight its planning robustness and recovery capability. The dataset and benchmark code will be released to support future research.
\end{abstract}

\section{Introduction}
\label{sec:intro}
End-to-end autonomous driving systems directly map raw sensor observations to actions within a unified framework \citep{jiang2023vad, zhou2024v2xpnp, zhou2025turbotrain, lei2025risk, xu2025wod,peng2026ctrlwam}. More recently, Vision-Language-Action (VLA) models have introduced the world knowledge and reasoning capabilities of Vision-Language Models (VLMs) into end-to-end driving \citep{ma2025dvlmadenhancediffusionvisionlanguagemodel,zhao2026bridgesim,zhou2026verifine}. Through unified perception, reasoning, and action generation, VLA models provide a promising foundation for generalizable driving.

Despite this progress, most existing VLAs \textbf{only learn from positive demonstrations, leading to limited robustness} \citep{yu2025survey}. However, real-world negative and recovery data, which capture negative behaviors that should be avoided, as well as recovery behavior from challenging scenarios, are often overlooked in datasets and models \citep{wang2024learning,xing2026drivevla}. Such negative-recovery data can provide targeted signals for efficient gains in model performance and robustness on long-tail scenarios, rather than relying on positive data concentrated in nominal driving, but the lack of real-world negative and recovery data limits progress in this domain \citep{liu2025takead, gao2026learning}. Besides, how to fully exploit those data is still an open question. First, they encode asymmetric supervision across positive, negative, and recovery behavior. Second, the challenging scenarios of negative and recovery samples often produce rollout groups dominated by similar suboptimal or infeasible trajectories, leading to uninformative advantages and unstable training.

Current VLA models also face \textbf{high action generation latency and large action experts}. Directly representing trajectories as discrete tokens provides a simple interface for jointly learning reasoning and planning \citep{kim2024openvla,zhou2025autovla}, but requires the large VLM backbone to decode every action autoregressively. The resulting latency grows linearly with the trajectory length, making high-frequency planning expensive. Continuous action experts provide a more efficient alternative \citep{li2025recogdrive,wang2025alpamayo}, yet existing designs commonly rely either on the final-layer VLM feature, which may omit intermediate visual and semantic information, or on dense KV caches from all VLM layers and a large action expert, incurring substantial memory overhead and latency.

To this end, we propose \textsc{\textbf{\gradientRGB{SpanVLA}{9,29,83}{23,99,180}}}, a robust and efficient VLA framework that learns from real-world negative and recovery samples with fast action bridging, as illustrated in \Cref{fig:framework}. First, we introduce \textbf{nuReasoning-NR}, a public extension of nuReasoning \citep{huang2026nureasoning} containing 36K reasoning-intensive scenarios, including 3K negative trajectories and 3K expert recovery trajectories. It is the first large-scale real-world driving dataset and benchmark with negative and recovery samples. Second, we introduce \textbf{Negative--Recovery Reinforcement Fine-Tuning (NR-RFT)} to exploit the asymmetric supervision provided by negative and recovery samples. Specifically, type-specific negative penalties suppress representative negative behaviors, while recovery rewards reinforce expert corrective actions. We additionally introduce an advantage correction to preserve informative optimization signals and encourage exploration toward feasible solutions. Third, \spanvla{} retains autoregressive reasoning and introduces an \textbf{Efficient Action Bridge} that aggregates multi-granular sparse KV caches to enable fast action generation. Instead of learning the flow directly from random noise \citep{li2025recogdrive}, it conditions on the historical trajectory and learns the transformation from past actions to future actions, improving generation quality and efficiency. 

Extensive evaluation on NAVSIM (v1 \& v2) \citep{dauner2024navsim, cao2025pseudo} and the nuReasoning \citep{huang2026nureasoning} benchmark demonstrates the state-of-the-art performance of our model. Empirical results validate that learning with real-world negative and recovery samples further improves planning performance and robustness. The main contributions are as follows:
\begin{enumerate}
    \item We propose \textbf{\spanvla{}}, a robust and efficient VLA framework that learns from negative-recovery samples with fast action bridging. It leverages a sparse KV cache with historical-trajectory-initialized flow matching to reduce action-generation latency.
    \item We introduce \textbf{nuReasoning-NR}, the first real-world driving dataset and benchmark containing both negative and expert recovery samples. It comprises 36K samples and enables systematic study of learning from real-world negative and recovery experiences.
    \item We introduce \textbf{NR-RFT}, a post-training method that exploits asymmetric negative-recovery supervision, together with an advantage correction for difficult rollout groups. 
    \item \spanvla{} achieves state-of-the-art performance across the NAVSIM v1, v2, and nuReasoning benchmarks, and learning with negative and recovery samples improves robustness.
\end{enumerate}

\section{Related Work}

\paragraph{VLA Models for Autonomous Driving.}
The reasoning capabilities and world knowledge of VLMs have motivated their application in autonomous driving \citep{hwang2024emma,cai2024driving,liu2026driveworld,zhang2026minddriver} and robotics \citep{kim2016fine, intelligence2025pi,liu2025robopilot}. Early approaches employ VLMs to provide high-level meta-actions \citep{xie2026latentvla,jiang2024senna,tian2024drivevlm} or auxiliary supervision \citep{pan2024vlp,xu2024vlm} for conventional planning systems, limiting end-to-end optimization. Recent VLA models instead unify reasoning and planning by representing trajectories as text \citep{rowe2025poutine, zhou2025opendrivevla, xing2025openemma, mao2023gpt} and discrete action or world tokens \citep{zhou2025autovla, zeng2025futuresightdrive}, but autoregressive action decoding incurs latency that grows with trajectory length. Thus, \spanvla{} generates continuous actions efficiently through a lightweight action expert and further fine-tunes the model with real-world negative-recovery samples to improve robustness.

\vspace{-0.1cm}
\paragraph{Driving Reasoning and Negative--Recovery Datasets.}
Large-scale real-world driving datasets provide multimodal observations and expert trajectories \citep{karnchanachari2024towards,openscene2023,xu2025wod}. Recent datasets further introduce driving question answering and reasoning annotations \citep{sima2024drivelm,wang2024omnidrive,park2025nuplanqa,arai2025covla,huang2026nureasoning}, but remain centered on positive driving data. Takeover-based approaches collect negative or corrective behaviors \citep{liu2025takead,fang2025corevla}, yet primarily rely on simulation, and real-world negative-recovery data remains a gap. In contrast, \textbf{nuReasoning-NR} extends nuReasoning with real-world negative trajectories and corresponding expert recovery trajectories, providing the first large-scale driving negative and recovery dataset.

\vspace{-0.1cm}
\paragraph{Action Bridging for VLA Models.}
Action experts bridge VLM representations with continuous control while avoiding autoregressive trajectory decoding \citep{wang2025alpamayo,li2025drivevla,dang2026drivefine,li2025recogdrive}. ReCogDrive \citep{li2025recogdrive} conditions a diffusion planner on the final-layer VLM representation, whereas Alpamayo \citep{wang2025alpamayo} uses dense KV caches from all VLM layers with a flow-matching planner. The former may omit intermediate visual and semantic information, while the latter introduces substantial memory and computational overhead. In contrast, \spanvla{} aggregates multi-granular KV caches from sparse VLM layers and employs a lightweight flow-matching expert initialized from the historical trajectory rather than pure noise.

\section{nuReasoning-NR Dataset and Benchmark}
We introduce \textbf{nuReasoning-NR}, the first real-world driving dataset and benchmark for learning from both negative and recovery samples. It extends nuReasoning \citep{huang2026nureasoning} and contains 36K samples: 30K positive reasoning samples, 3K negative samples, and 3K recovery samples. All three subsets cover the same categories of interaction-intensive driving scenarios, as shown in \Cref{fig:data_distribution}.

\vspace{-0.1cm}
\paragraph{Real-World Negative and Recovery Data.}
We collect negative and recovery samples from early-stage exploratory driving tests. A negative segment records suboptimal ego behavior, while its recovery counterpart records the subsequent expert correction. Each segment contains its own observation history and reference trajectory as a data sample. 

\vspace{-0.1cm}
These samples provide distinct forms of supervision. A negative trajectory identifies behavior that should be suppressed, including inefficient or hesitant maneuvers as well as potentially unsafe actions. An expert recovery trajectory demonstrates how to correct behavior from a difficult driving state. Their event-level correspondence captures both the suboptimal behavior and the subsequent correction, while their different reference semantics motivate the asymmetric supervision in NR-RFT. Representative examples are shown in \Cref{fig:data_distribution}.

\vspace{-0.1cm}
\paragraph{Positive Reasoning Data.}
The positive subset provides expert trajectories and reasoning annotations for supervised fine-tuning. We generate reasoning annotations using an automated pipeline based on Gemini-3-Pro \citep{gemini3report2025}. The pipeline identifies decision-relevant scene elements, analyzes their states, selects longitudinal and lateral actions, and summarizes the resulting decisions into concise reasoning traces. Automated checks assess annotation-format completeness and alignment with the reference actions, followed by a human quality audit. Further annotation details are provided in Appendix~\ref{app:nureasoning_details}.

\vspace{-0.1cm}
\paragraph{nuReasoning-NR Benchmark.}
Within the 36K samples, we designate 300 negative scenarios and 300 recovery scenarios for benchmark evaluation. The negative subset evaluates planning in situations associated with suboptimal recorded behavior, while the recovery subset evaluates corrective planning from difficult states. To support the evaluation, we adapt the NAVSIM evaluation pipeline \citep{dauner2024navsim} using the dataset's object annotations and HD maps. This benchmark enables systematic evaluation of planning performance across both scenario types and of the benefits of negative-recovery supervision. Additional dataset details are provided in Appendix~\ref{app:nureasoning_details}.

\section{SpanVLA}

\begin{figure}[t]
    \centering
    \includegraphics[width=\linewidth]{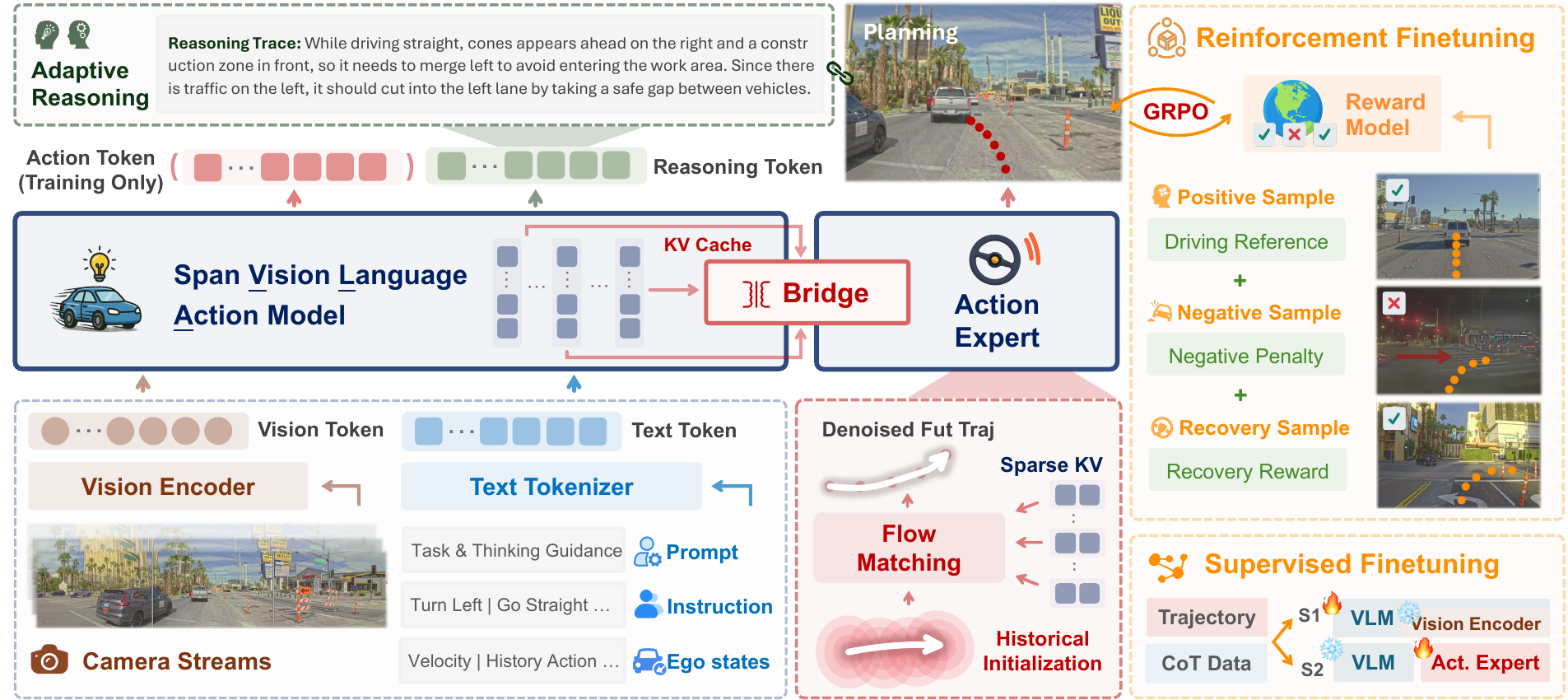}
    \caption{\textbf{Overview of \spanvla{}.} Supervised fine-tuning aligns reasoning and planning, while NR-RFT uses real-world negative and recovery samples to penalize undesirable behaviors and reward expert corrections. NR-RFT updates the policy using corrected group-relative advantages. For efficient inference, a lightweight action bridge generates continuous trajectories through flow matching conditioned on sparse-layer KV caches and initialized from the historical trajectory.}
    \label{fig:framework}
    \vspace{-0.15cm}
\end{figure}

\spanvla{} combines a VLM backbone for reasoning with a lightweight action bridge for continuous trajectory generation, as illustrated in \Cref{fig:framework}. Training uses supervised fine-tuning (SFT) and Negative--Recovery Reinforcement Fine-Tuning (NR-RFT).

\subsection{VLM Backbone}
\paragraph{Model Inputs.} The backbone takes a multimodal query $q=(\mathcal{V}^t,\mathcal{T}^t)$ at driving time $t$. The visual input $\mathcal{V}^t$ contains four frames per camera at 2\,Hz. The textual input $\mathcal{T}^t$ includes a task prompt, a driving instruction, and ego states comprising acceleration, velocities, and the historical trajectory.

\vspace{-0.1cm}
\paragraph{Reasoning with Autoregressive Decoding.} 
During training, to enable the model to learn how to reason for planning, we introduce an additional discrete action generation task following reasoning in the VLM, which unifies reasoning and planning within the SFT, as follows: 
\begin{equation}
    [\mathcal{T}_{\text{Reason}}, (A_{\text{token}})] = \mathrm{VLM}(\mathcal{V}^t, \mathcal{T}^t); A_{\text{token}}=\{a_{\text{token}}^\tau\}^{t+T_f}_{\tau=t},
    \label{vlm_backbone}
\end{equation}  
where we leverage the action codebook of AutoVLA \citep{zhou2025autovla} to discretize the trajectories into action tokens $A_{\text{token}}$. We further adopt adaptive reasoning: slow thinking generates reasoning before actions, while fast thinking directly generates actions. At inference, the VLM emits an action-start token and delegates continuous trajectory generation to the action bridge.

\begin{figure}[t]
    \centering
    \includegraphics[width=\linewidth]{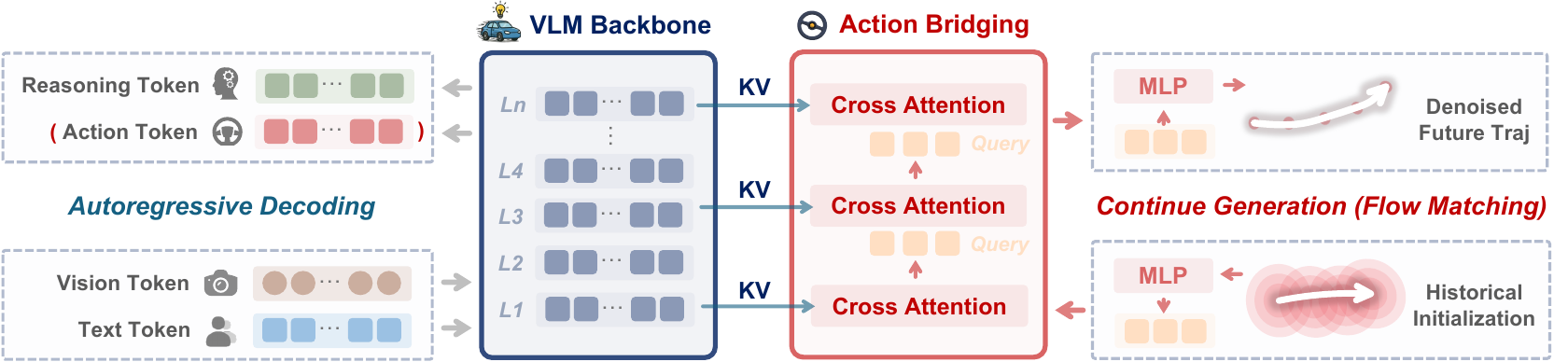}
    \caption{\textbf{Efficient action bridging in \spanvla{}.} The VLM backbone generates reasoning autoregressively. A lightweight action bridge then generates a continuous trajectory through flow matching, conditioned on sparse KV caches from selected VLM layers and initialized from the historical trajectory, to accelerate action generation.}
    \label{fig:bridge}
    \vspace{-0.15cm}
\end{figure}

\subsection{Efficient Action Bridge}
\label{sec:action_bridge}

\paragraph{Sparse KV-Cache Conditioning.}
The action bridge uses a lightweight Transformer $f_\phi$ conditioned on KV caches from selected VLM layers, denoted by $\mathbf{c}_{\mathrm{vlm}}$. These caches encode visual observations, textual inputs, and the generated reasoning. Accessing representations across selected depths provides multi-granular context while reducing the computation associated with dense layer-wise conditioning. The bridge retains the VLM's number of attention heads with a smaller hidden dimension. At each flow step, it attends to the selected caches to predict a vector field $\mathbf{v}_{\tau}$ from the current trajectory state and a flow-time embedding, as illustrated in \Cref{fig:bridge}.

\paragraph{Historical-Trajectory Initialization.}
Rather than starting from pure Gaussian noise $\mathcal{N}(0, \mathbf{I})$ \citep{wang2025alpamayo}, our method initializes from the historical trajectory embedding $\mathbf{a}_{\text{his}}$ with MLP layers and directly learns the transition from the historical action space to the future action space: $\mathbf{a}_{\text{his}} \xrightarrow{\mathbf{v}_{\tau}} \mathbf{a}$, improving performance and sampling efficiency. Moreover, to enhance robustness, we inject some Gaussian noise into the historical trajectory embeddings during training. In practice, we employ the optimal transport displacement map \citep{lipman2022flow}. For training, we sample $\mathbf{a}_{\tau}=\tau\mathbf{a}+(1-\tau)\mathbf{a}_{\text{his}}$. In inference, the flow matching denoising with in each $\Delta \tau$ step is:
\begin{equation}
\mathbf{a}_{\tau+\Delta\tau}
= \mathbf{a}_{\tau}
+ \Delta\tau\,f_{\phi}
(\mathbf{a}_{\tau},\tau,\mathbf{c}_{\mathrm{vlm}}).
\end{equation}

\subsection{Supervised Fine-Tuning}

\vspace{-0.1cm}
\paragraph{VLM Backbone.} We supervise the VLM with expert trajectories and reasoning annotations. Slow-thinking targets contain reasoning followed by action tokens, while fast-thinking targets contain only action tokens. All training data are curated with ground-truth assistant responses to ensure high-quality supervision for the VLM. Thus, the loss function for the output sequence $o=[\mathcal{T}_{\text{Reason}}, A_{\text{token}}]$ is defined as:
\begin{equation}
\resizebox{1.0\textwidth}{!}{%
$
\mathcal{L}_{\text{LM}} = - \frac{1}{N}\sum\limits_{i=1}^{N} \log p_\theta(o_i \mid o_{<i}, \mathcal{V}^t, \mathcal{T}^t), \mathcal{L}_{\text{Action}} = - \frac{1}{T}\sum\limits_{i=L+1}^{L+T} \log p_\theta(o_i \mid o_{<i}, \mathcal{V}^t, \mathcal{T}^t)
$,
}
\end{equation}
where $N = L + T$, and $p_\theta$ denotes the predicted distribution with policy $\theta$. The final loss is $ \mathcal{L}_i^\text{SFT} = w_i \cdot \mathcal{L}_{\text{LM},i} + \mathcal{L}_{\text{Action},}$, where $w_i=0$ for slow-thinking, and $w_i=1$ when CoT is activated.

\vspace{0.1cm}
\paragraph{Action Bridging.} It is trained with a conditional flow matching loss, which encourages the model to learn the optimal transport path from the historical action to the future planning action:
\begin{equation}
    \mathcal{L}_{\text{FM}} = \mathbb{E}_{\tau \in [0,1], \mathbf{c}_{\text{vlm}}\in[\mathcal{V}^t, \mathcal{T}^t, \mathcal{T}_{\text{Reason}}]} \| f_\phi(\mathbf{a}_\tau, \tau, \mathbf{c}_{\text{vlm}}) - \mathbf{v}_\tau(\mathbf{a}_\tau) \|^2,
\end{equation}
where $\tau$ is sampled by shifted normal distribution during training, and the target vector field is $\mathbf{v}_{\tau}(\mathbf{a}_{\tau})=\frac{d\mathbf{a}_{\tau}}{d\tau}=\mathbf{a}-\mathbf{a}_{\text{his}}$. Moreover, following Alpamayo \citep{wang2025alpamayo}, we stop the gradients between the VLM backbone and the bridging module to preserve their learned knowledge. In practice, we first train the VLM backbone and then fine-tune the action bridge separately.

\subsection{Negative--Recovery Reinforcement Fine-Tuning}
\label{sec:nr_rft}

NR-RFT adapts group relative policy optimization (GRPO)~\citep{shao2024deepseekmath,jiang2025alphadrive} to the asymmetric supervision of negative and recovery samples, with an advantage correction for difficult rollout groups to encourage exploration. For each query $q$, we sample $G$ VLM responses and decode their action tokens into trajectories $o_i=\{\mathbf{p}_{i,t}\}_{t=0}^{T}$. Rewards evaluate each trajectory and its associated reasoning, and update the VLM token policy.

\vspace{-0.1cm}
\paragraph{Asymmetric Behavioral Rewards.}
Given the reference trajectory $o^{\mathrm{ref}}=\{\mathbf{p}^{\mathrm{ref}}_t\}_{t=0}^{T}$, we define a bounded reference-matching term:
\begin{equation}
    d(o_i,o^{\mathrm{ref}})
    =\frac{1}{T+1}\sum_{t=0}^{T}
    \|\mathbf{p}_{i,t}-\mathbf{p}^{\mathrm{ref}}_t\|_2,
    \qquad
    r_{\text{Ref}}(q,o_i)
    =\operatorname{clip}\!\left(
        1-\frac{d(o_i,o^{\mathrm{ref}})}{\delta},0,1
    \right),
    \label{eq:reference_reward}
\end{equation}
where $\delta>0$ is a proximity threshold. We set $r_{\text{Negative},i}=r_{\text{Ref}}(q,o_i)$ for negative samples and $r_{\text{Recover},i}=r_{\text{Ref}}(q,o_i)$ for recovery samples; each term is zero for other types. The final reward is

\begin{equation}
    r_i
    =r_{\text{Driving},i}
    -\lambda_{\text{N}}r_{\text{Negative},i}
    +\lambda_{\text{R}}r_{\text{Recover},i}
    -\lambda_{\text{C}}r_{\text{CoT},i},
    \label{eq:behavioral_rewards}
\end{equation}

where the non-negative weights control the corresponding terms. This design penalizes proximity to undesirable behavior and rewards agreement with expert corrections. The reference-matching term vanishes beyond $\delta$, so further deviation from a negative reference provides no additional benefit.

The driving reward uses PDMS or EPDMS according to the evaluation setting~\citep{dauner2024navsim,cao2025pseudo}. For negative samples, we omit ego progress because the recorded trajectory is an undesirable reference. The reasoning cost $r_{\text{CoT},i}$ penalizes unnecessarily long reasoning and is replaced by a fixed penalty when a rule-based check detects disagreement between reasoning and the predicted maneuver. Detailed reward definitions are provided in the supplementary material.

\vspace{-0.1cm}
\paragraph{Advantage Correction.}
Challenging negative and recovery scenarios can produce rollout groups with nearly identical rewards or all-zero rewards. Standard GRPO uses
$A_i^{\mathrm{GRPO}}=(r_i-\mu)/(\sigma+\varepsilon_{\mathrm{num}})$,
where $\mu$ and $\sigma$ are the group reward mean and standard deviation. Standard-deviation normalization can amplify small reward differences, while groups with identical final rewards yield zero advantages.

We remove standard-deviation normalization to preserve the original reward differences. For groups with all-zero rewards, we additionally include a virtual reward of $1$ in the calculation. Let $F=\mathbb{I}[r_{\text{Driving},1}=\cdots=r_{\text{Driving},G}=0]$, while $r_i$ remains the final reward defined in \eqref{eq:behavioral_rewards}. Then, the corrected advantage is
\begin{equation}
    \widetilde{\mu}
    =\frac{\sum_{j=1}^{G}r_j+F}{G+F},
    \qquad
    A_i=r_i-\widetilde{\mu}.
    \label{eq:nr_advantage}
\end{equation}
When $F=1$ and the final rewards are also all zero, every sampled output receives $A_i=-1/(G+1)$, providing a penalty signal on the sampled zero-reward outputs. The virtual reward affects only the baseline and contributes no additional output to the policy objective. When $F=0$, $A_i=r_i-\mu$.

\section{Experiments}

\subsection{Experimental Setup}

\paragraph{Dataset.}
Training data are drawn from the NAVSIM \texttt{navtrain} split (100K scenarios)~\citep{dauner2024navsim}, nuReasoning, and the training portion of nuReasoning-NR. Expert trajectories and reasoning annotations provide supervised training signals, while negative and recovery samples introduce supervision on undesirable behaviors and expert corrections during NR-RFT.

\vspace{-0.1cm}
\paragraph{Benchmark.}
We evaluate on NAVSIM v1 \texttt{navtest} using PDMS and NAVSIM v2 \texttt{navtest} and \texttt{navhard} using EPDMS~\citep{dauner2024navsim,cao2025pseudo}. The \texttt{navhard} split further assesses planning in challenging scenarios through pseudo closed-loop simulation. On nuReasoning~\citep{huang2026nureasoning}, we report NPS and ADE to evaluate planning in interaction-intensive scenarios.
On nuReasoning-NR, we report PDMS separately for the 300 negative and 300 recovery scenarios, together with their average and component metrics.

\begin{table}[t]
    \caption{\footnotesize Comparison on the \textbf{nuReasoning-NR benchmark}. Results are reported in PDMS.}
  \vspace{0.1cm}
  \renewcommand{\arraystretch}{1.1}
  \label{tab:negative_recovery}
  \centering
  \scriptsize
  \setlength{\tabcolsep}{5pt}
  \begin{tabular}{l|cc|c|ccccc}
    \toprule
    Methods
    & Negative $\uparrow$
    & Recovery $\uparrow$
    & Average PDMS $\uparrow$
    & NC $\uparrow$
    & DAC $\uparrow$
    & EP $\uparrow$
    & TTC $\uparrow$
    & Comf. $\uparrow$ \\
    \midrule

    \rowcolor[HTML]{eaf1f8}
    \multicolumn{9}{l}{\textit{Models Trained with Positive Data Only}} \\
    LTF \citep{chitta2022transfuser}
    & 62.7 & 51.0 & 56.8
    & 99.2 & 59.9 & 55.7 & 99.8 & 89.4 \\
    
    ReCogDrive \citep{li2025recogdrive}
    & 64.0 & 54.0 & 59.0
    & 99.6 & 61.8 & 58.0 & 99.7  & 90.9 \\

    AutoVLA \citep{zhou2025autovla}
    & 64.2 & 54.0 & 59.1
    &  99.4 & 62.2 & 58.0 & 99.7 & 90.4 \\
    \spanvla{}
    & 65.1 & 56.1 & 60.6
    & 99.9 & 62.3 & 58.6 & 100.0 & 92.5 \\
    
    \midrule
    \rowcolor[HTML]{fff3e9}
    \multicolumn{9}{l}{\textit{Models Trained with Additional NR Data}} \\
    AutoVLA + NR
    & 68.8 & 59.5 & 64.2
    & 99.7 & 67.3 & 62.5 & 99.8 & 90.5 \\
    \spanvla{} + NR
    & \textbf{74.1}
    & \textbf{65.1}
    & \textbf{69.6}
    & \textbf{100.0} & \textbf{71.2} & \textbf{66.9} & \textbf{100.0} & \textbf{93.2} \\
    
    \bottomrule
  \end{tabular}
  \vspace{-0.3cm}
\end{table}

\begin{table}[t]
  \caption{\footnotesize Comparison on the \textbf{nuReasoning benchmark}. Results are reported using planning metrics.}
  \vspace{0.1cm}
  \renewcommand{\arraystretch}{1.1}
  \label{tab:nuReasoning}
  \centering
  \scriptsize
  \setlength{\tabcolsep}{11pt}

  \begin{tabular}{l|ccccc|cc}
    \toprule
    \multirow{1}{*}{Methods}
    & NC $\uparrow$
    & DA $\uparrow$
    & EP $\uparrow$
    & CF $\uparrow$
    & HL $\uparrow$
    & NPS $\uparrow$
    & ADE $\downarrow$ \\
    \midrule

    UniAD \citep{hu2023planning}
    & 88.9 & 87.6 & 89.6 & 92.6 & 48.8 & 55.7 & 2.1 \\

    DiffusionDrive \citep{liao2024diffusiondrive}
    & 90.2 & 88.3 & 90.5 & 95.0 & 52.0 & 57.9 & 1.9 \\

    AutoVLA \citep{zhou2025autovla}
    & 90.9 & 86.5 & 89.3 & \textbf{99.9} & 49.9 & 59.1 & 2.1 \\

    Alpamayo-1.5 \citep{wang2025alpamayo}
    & 90.3 & 86.1 & 86.5 & 97.9 & 33.8 & 50.5 & 2.9 \\

    nuVLA \citep{huang2026nureasoning}
    & 96.3 & 96.3 & \textbf{90.7} & 99.7 & 62.2 & 73.1 & 1.6 \\

    \midrule
    \rowcolor[HTML]{fff3e9}
    \spanvla{} (Post-RFT)
    & \textbf{96.5} & \textbf{96.4} & 90.3  & \textbf{99.9} &  \textbf{63.4} & \textbf{73.3} & \textbf{1.4} \\

    \bottomrule
  \end{tabular}

  \vspace{-0.3cm}
\end{table}

\vspace{-0.1cm}
\paragraph{Implementation Details.} We use the Qwen2.5-VL-3B model \citep{bai2025qwen2} as the VLM backbone. For SFT, we employ Fully Sharded Data Parallel (FSDP) training on 8 NVIDIA H200 GPUs. The per-GPU batch size is set to 4, with a gradient accumulation step of 4. For the action bridge, the batch size is set to 16, and we selected the feature from the VLM with an interval of two. For NR-RFT, we adopt LoRA adapters~\citep{hu2022lora} for efficient fine-tuning. The learning rate for NR-RFT is set to $3 \times 10^{-5}$ and the group sample size is $64$. We perform a single policy update per step, resulting in a simplified objective that eliminates the need for clipping or maintaining an old policy. Additional implementation details are provided in the Appendix.

\vspace{-0.1cm}
\subsection{Main Results}
This section reports the main results of \spanvla{}, and more results are provided in the Appendix.

\vspace{-0.1cm}
\paragraph{nuReasoning-NR Benchmark.}
\Cref{tab:negative_recovery} evaluates whether real-world negative and recovery supervision improves planning in challenging scenarios. Incorporating NR supervision increases the average PDMS of \spanvla{} by 14.9\%. The gains occur on both subsets: negative scenarios improve by 13.8\%, and recovery scenarios improve by 16\%. The benefit extends to AutoVLA, whose average PDMS increases from 59.1 to 64.2 with NR training. This supports the value of negative-recovery supervision beyond the \spanvla{} architecture. 

\begin{table*}[t]
  \centering
  \caption{\footnotesize
  Comparison with SOTA methods on the \texttt{navtest} benchmarks of
  \textbf{NAVSIM v1} and \textbf{NAVSIM v2}, and on the
  \texttt{navhard} benchmark of \textbf{NAVSIM v2}.
  NAVSIM v1 is evaluated using PDMS, while NAVSIM v2 is evaluated
  using EPDMS. * denotes best-of-N selection with N = 6, where the candidate with the highest score is chosen. Detailed metric breakdowns are provided in
  Appendix~\Cref{tab:navsim_v1,tab:navsim_v2,tab:navhard}.}
  \label{tab:navsim_main_results}
  \vspace{0.1cm}

  \begin{minipage}[c]{0.59\textwidth}
    \centering
    \scriptsize
    \setlength{\tabcolsep}{3pt}
    \renewcommand{\arraystretch}{1.1}

    \begin{tabularx}{\linewidth}{
      >{\raggedright\arraybackslash}X|cc|c|c
    }
      \toprule
      \multirow{2}{*}{Methods}
      & \multirow{2}{*}{Cam.}
      & \multirow{2}{*}{Lid.}
      & \multicolumn{1}{c|}{\textbf{NAVSIM v1}}
      & \multicolumn{1}{c}{\textbf{NAVSIM v2}} \\
      & & & PDMS $\uparrow$ & EPDMS $\uparrow$ \\
      \midrule

      \rowcolor[HTML]{eaf1f8}
      \multicolumn{5}{l}{\textit{Conventional End-to-end-based Methods}} \\
      TransFuser \citep{chitta2022transfuser}
          & \checkmark & \checkmark & 84.0 & 76.7 \\
      Hydra-MDP \citep{li2024hydra}
          & \checkmark & \checkmark & 86.5 & 81.4 \\
      DiffusionDrive \citep{liao2024diffusiondrive}
          & \checkmark & \checkmark & 88.1 & 84.5 \\

      \midrule
      \rowcolor[HTML]{eaf1f8}
      \multicolumn{5}{l}{\textit{VLA-Based Methods}} \\
      AutoVLA \citep{zhou2025autovla}
          & \checkmark & -- & 89.1 & 82.4 \\
      ReCogDrive \citep{li2025recogdrive}
          & \checkmark & -- & 89.6 & 83.6 \\
      DriveVLA-W0 \citep{li2025drivevla}
          & \checkmark & -- & 90.2 & 86.1 \\
      Qwen-Drive \citep{zhou2026qwen}
          & \checkmark & -- & 90.7 & --- \\

      \midrule
      \rowcolor[HTML]{fff3e9}
      \multicolumn{5}{l}{\textit{\textbf{Ours}}} \\
      \spanvla{} (One-shot)
          & \checkmark & -- & 82.1 & 79.4 \\
      \spanvla{} (Post-RFT, positive only)
          & \checkmark & -- & 90.8 & 86.9 \\
      \spanvla{}*
          & \checkmark & -- & \textbf{93.3} & \textbf{90.5} \\

      \bottomrule
    \end{tabularx}
  \end{minipage}%
  \hfill
  \begin{minipage}[c]{0.38\textwidth}
    \centering
    \scriptsize
    \setlength{\tabcolsep}{3pt}
    \renewcommand{\arraystretch}{1.14}

    \begin{tabularx}{\linewidth}{
      >{\raggedright\arraybackslash}X|c
    }
      \toprule
      \multirow{2}{*}{Methods}
      & \multicolumn{1}{c}{\textbf{v2 navhard}} \\
      & EPDMS $\uparrow$ \\
      \midrule

      \rowcolor[HTML]{eaf1f8}
      \multicolumn{2}{l}{\textit{Conventional End-to-end-based Methods}} \\
      TransFuser \citep{chitta2022transfuser} 
          & 23.1 \\
      DiffusionDrive \citep{liao2024diffusiondrive}
          & 24.2 \\
      Senna-E2E \citep{jiang2026senna}
          & 27.2 \\
      MindDrive \citep{sun2025minddrive}
          & 30.9 \\
      World4Drive \citep{zheng2025world4drive}
          & 34.9 \\
      Navformer \citep{wang2024navformer}
          & 34.1 \\
      iPad \citep{guo2026ipad}
          & 34.7 \\
      RAP-DINO \citep{feng2025rap}
          & 36.9 \\

      \midrule
      \rowcolor[HTML]{fff3e9}
      \multicolumn{2}{l}{\textit{\textbf{Ours}}} \\
      \spanvla{} (Post-RFT, positive only)
          & 40.1 \\
      \spanvla{} (Post-RFT, with NR)
          & 40.8 \\
      \spanvla{}*
          & \textbf{47.9} \\

      \bottomrule
    \end{tabularx}
  \end{minipage}

  \vspace{-0.3cm}
\end{table*}

\vspace{-0.1cm}
\paragraph{NAVSIM Benchmark.}
\Cref{tab:navsim_main_results} compares \spanvla{} with state-of-the-art (SOTA) conventional end-to-end and VLA-based methods on the \texttt{navtest} benchmarks of NAVSIM v1 and NAVSIM v2, as well as the NAVSIM v2 \texttt{navhard} benchmark. RFT substantially improves the one-shot policy, while incorporating negative and recovery data provides further gains over positive-only RFT on challenging \texttt{navhard}. \spanvla{} achieves the best overall performance across all three benchmarks. Detailed metric breakdowns are provided in Appendix~\Cref{tab:navsim_v1,tab:navsim_v2,tab:navhard}.

\vspace{-0.1cm}
\paragraph{nuReasoning Benchmark.}
On nuReasoning, \spanvla{} achieves the highest NPS and lowest ADE among the methods
in \Cref{tab:nuReasoning}. These results complement the negative-recovery evaluation by demonstrating strong planning performance across interaction-intensive driving scenarios.

\vspace{-0.1cm}
\paragraph{Efficient Action Bridging.}
\Cref{tab:action_policy} compares various action policies. For 50-waypoint trajectories, the bridge reduces action-generation latency, achieving a $21.5\times$ speedup. The reported total latency has a reduction of approximately $74\%$. Increasing the output from 10 to 50 waypoints leaves the bridge's measured generation time approximately unchanged, supporting efficient trajectory generation at high temporal resolution. The $L_1$ head is faster at 0.02\,s, but achieves 85.1 PDMS compared with 90.8 for flow matching in the reported 10-waypoint evaluation. The proposed bridge thus provides a favorable balance between planning quality and action-generation efficiency.

\begin{table*}[t]
    \centering
    \setlength{\tabcolsep}{3.4pt}
    \caption{\footnotesize Comparison for action policy on the \textbf{NAVSIM v1}. The fixed VLM encoding + prefilling time is with the same inputs, and reasoning tokens are the average tokens across 10K samples. FM: flow-matching.}
    \vspace{0.1cm}
    \label{tab:action_policy}
    \scriptsize
    \setlength{\tabcolsep}{7.5pt} 
    \renewcommand{\arraystretch}{1.08} 
    \begin{tabular}{l | c | c | c c c c}
        \toprule
        Methods & Action & PDMS $\uparrow$ & \makecell[c]{VLM Encoding \\ + Prefilling {(s)} $\downarrow$}  & \makecell[c]{Reasoning \\ Generation {(s)}$\downarrow$} & \makecell[c]{Trajectory \\ Generation (s)$\downarrow$} & Total (s) $\downarrow$ \\
        \midrule
        \rowcolor[HTML]{eaf1f8} \multicolumn{7}{l}{\textit{Autoregressive Decoding}} \\
        \multirow{2}{*}{AutoVLA \citep{zhou2025autovla}} & 10 & 89.1 & 0.09 & 0.76 {\tiny (23 tokens)} & 0.40 {\tiny (12 tokens)} & 1.25 \\
                                                       & 50 & -- & 0.09  & 0.76 {\tiny (23 tokens)} & 1.72 {\tiny (52 tokens)}  & 2.57   \\
        \midrule
        \rowcolor[HTML]{fff3e9} \multicolumn{7}{l}{\textit{Action Expert}} \\
        \multirow{2}{*}{\spanvla{} ($L_1$ Head)}              & 10 & 85.1 & 0.09 & 0.50 {\tiny (15 tokens)} & 0.02  & 0.61 \\
                                                       & 50 & --   & 0.09 & 0.50 {\tiny (15 tokens)}   & 0.02  & 0.61   \\
        \midrule 
        \multirow{2}{*}{\spanvla{} (FM)}        & 10 & \textbf{90.8} & 0.09 & 0.50 {\tiny (15 tokens)}  & 0.08 {\tiny (5 steps)}  & 0.67 \tiny{\textcolor{darkgreen}{-46\%}}  \\
                                                       & 50 & --  & 0.09 & 0.50 {\tiny (15 tokens)}   & 0.08 {\tiny (5 steps)}   & 0.67  \tiny{\textcolor{darkgreen}{-74\%}}   \\
        \bottomrule
    \end{tabular}
\end{table*}

\begin{table}[t]
  \caption{Comparison for \spanvla{} with different bridging layers on \textbf{NAVSIM v1}. Full Caching: Following Alpamayo \citep{wang2025alpamayo}, an empty VLM decoding module of the backbone for bridging. ``Last Layer,'' ``Interval 4,'' and ``Interval 2'' denote extracting features from the final layer, every four layers, and every two layers. HI: historical initialization.}
  \vspace{0.1cm}
  \renewcommand{\arraystretch}{1.1}
  \label{tab:ablation_bridge}
  \centering
  \scriptsize
  \setlength{\tabcolsep}{15pt}
  \begin{tabular}{l|c|cc|c >{\columncolor[HTML]{fff3e9}}c}
    \toprule
    Metrics & Full Caching & Last Layer & Interval 4  & Interval 2 w/o HI & \textbf{Interval 2} \\
    \midrule
    PDMS $\uparrow$      & 88.1     &79.3  & 82.2  & 86.4      & \textbf{90.8}      \\
    Traj. Gen. (s) $\downarrow$  & 0.18    & \textbf{0.01}  & 0.05  & 0.08      & 0.08      \\
    \bottomrule
  \end{tabular}
  \vspace{-0.1cm}
\end{table}

\subsection{Ablation Studies for NR-RFT}

\begin{wrapfigure}{r}{0.48\textwidth}
  \vspace{-0.3cm}
  \centering
  \includegraphics[width=\linewidth]{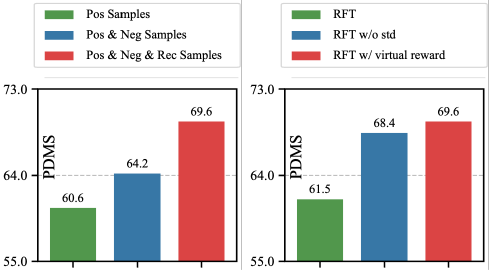}
  \caption{\footnotesize \textbf{Ablations of NR-RFT on nuReasoning-NR.} \textbf{Left:} Positive-only, positive--negative, and positive--negative--recovery supervision. \textbf{Right:} Standard GRPO, mean-centered advantages without standard-deviation normalization, and the full estimator with virtual-reference correction.}
  \label{fig:abl_nr}
  \vspace{-0.8cm}
\end{wrapfigure}

\vspace{-0.1cm}
\paragraph{Negative and Recovery Supervision.}
\Cref{fig:abl_nr} (left) compares training with positive samples, 
\noindent 
positive and negative samples, and all three sample types. Adding negative supervision improves PDMS on nuReasoning-NR from 60.6 to 64.2, while incorporating recovery supervision further increases it to 69.6. The additional gain from recovery samples supports their value beyond negative supervision alone. Together, these results support learning both which behaviors to avoid and how to correct behavior from difficult driving states.

\vspace{-0.1cm}
\paragraph{Advantage Correction.}
\Cref{fig:abl_nr} (right) evaluates the two components of advantage correction. Removing standard-deviation normalization improves PDMS from 61.5 to 68.4.
Adding the virtual reference for groups with all-zero final rewards further increases PDMS to 69.6. These results support their effectiveness.

\vspace{-0.1cm}
\subsection{Ablation Studies for Action Bridging}
\vspace{-0.1cm}
\paragraph{Effect of Historical Initialization.} 
\Cref{tab:ablation_bridge} isolates historical initialization under the same interval-2 cache configuration. Historical initialization improves PDMS from 86.4 to 90.8 while maintaining a trajectory-generation latency of 0.08\,s. This shows that past motion provides useful guidance for future trajectory generation without increasing the measured inference time.

\vspace{-0.1cm}
\paragraph{Sparse KV-Cache Conditioning.}
Conditioning on every second VLM layer achieves 90.8 PDMS, compared with 82.2 using every fourth layer and 79.3 using only the final layer. These results support accessing representations across VLM depths, while showing that overly sparse conditioning degrades planning. The full-caching baseline uses a larger bridge without
historical initialization, achieves 88.1 PDMS with 0.18\,s latency. Our configuration achieves higher planning performance at 0.08\,s, providing an effective balance between contextual information and computation.

\begin{figure}[t]
    \centering
    \vspace{-0.3cm}
    \includegraphics[width=0.98\linewidth]{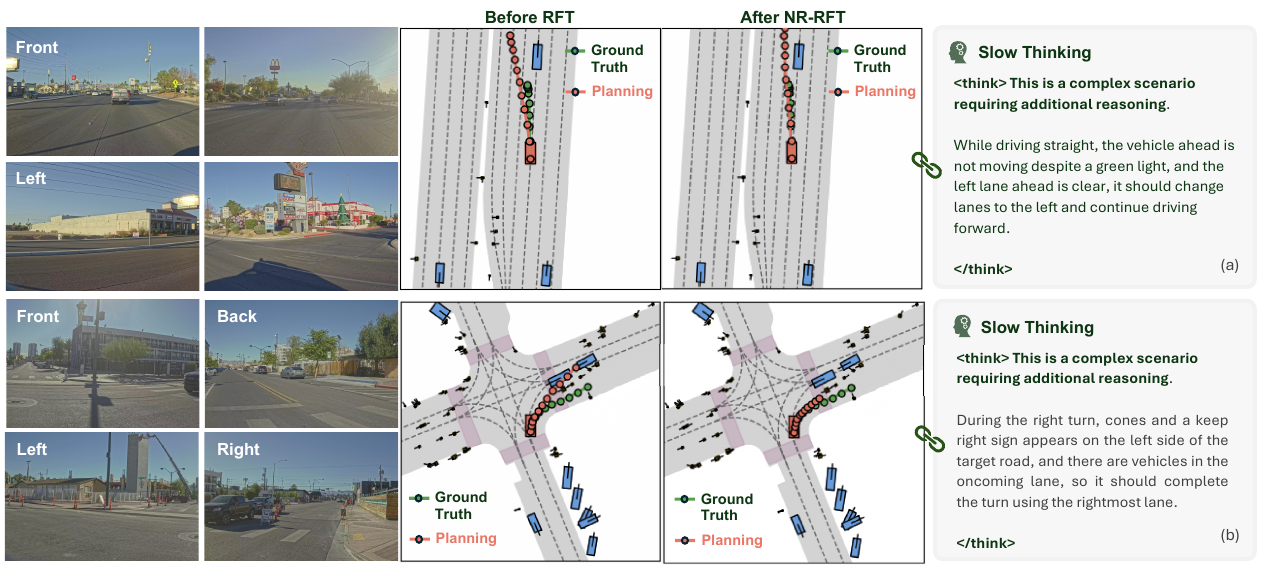}
    \vspace{-0.3cm}
    \caption{\footnotesize Comparison of planning and reasoning performance of \spanvla{} in \textbf{negative and recovery samples} of nuReasoning-NR before and after NR-RFT.}
    \label{fig:negative_vis}
    \vspace{-0.5cm}
\end{figure}

\vspace{-0.1cm}
\subsection{Qualitative Results}
\Cref{fig:data_distribution,fig:negative_vis} illustrate how NR-RFT improves planning in negative and recovery scenarios. In the negative scenarios, the updated policy predicts an earlier merge before a construction-zone lane closure (\Cref{fig:data_distribution}, upper left)
and a more decisive lane change around a stationary lead vehicle (\Cref{fig:negative_vis}(a)). In the recovery scenarios, it predicts a return to the original lane after passing an obstructing truck (\Cref{fig:data_distribution}, lower left) and a right turn aligned with the available lane through a constrained intersection (\Cref{fig:negative_vis}(b)). Additional examples are provided in Appendix~\ref{app:qualitative_results}.

\section{Conclusion}
\vspace{-0.1cm}
We presented \spanvla{}, a robust and efficient driving VLA, together with \textbf{nuReasoning-NR}, the first real-world driving dataset and benchmark for learning from negative and recovery samples.
NR-RFT exploits their asymmetric supervision by penalizing undesirable behaviors and rewarding expert corrections, with advantage correction for difficult rollout groups. For efficient action generation, \spanvla{} retains autoregressive reasoning and uses a lightweight flow-matching bridge conditioned
on sparse-layer KV caches and initialized from the historical trajectory. Experiments on NAVSIM v1/v2, nuReasoning, and nuReasoning-NR demonstrate strong planning performance, improved robustness in challenging scenarios, and efficient trajectory generation. These results highlight the value of real-world negative and recovery supervision for driving VLAs. The dataset and benchmark code will be released to support further research.




\bibliography{main}
\bibliographystyle{iclr2027_conference}

\clearpage
\setcounter{section}{0}
\setcounter{figure}{0}
\setcounter{table}{0}
\setcounter{equation}{0}
\renewcommand{\thesection}{\Alph{section}}
\renewcommand{\thefigure}{S\arabic{figure}}
\renewcommand{\thetable}{S\arabic{table}}
\renewcommand{\theequation}{S\arabic{equation}}

\title{\texorpdfstring{\raisebox{-.18\height}{\includegraphics[width=0.07\textwidth]{figure/logo.png}} \textit{SpanVLA}}{SpanVLA} Supplementary Material}

\let\oldaddcontentsline\addcontentsline
\let\oldaddtocontents\addtocontents

\renewcommand{\addcontentsline}[3]{}
\renewcommand{\addtocontents}[2]{}

\begingroup
\author{}
\maketitle
\endgroup

\let\addcontentsline\oldaddcontentsline
\let\addtocontents\oldaddtocontents

\vspace{0.75cm}
\makeatletter
\let\oldaddcontentsline\addcontentsline
\renewcommand{\addcontentsline}[3]{%
    \def\@tempa{#1}\def\@tempb{toc}%
    \ifx\@tempa\@tempb
        \oldaddcontentsline{stc}{#2}{#3}%
    \else
        \oldaddcontentsline{#1}{#2}{#3}%
    \fi
}
\let\oldaddtocontents\addtocontents
\renewcommand{\addtocontents}[2]{%
    \def\@tempa{#1}\def\@tempb{toc}%
    \ifx\@tempa\@tempb
        \oldaddtocontents{stc}{#2}%
    \else
        \oldaddtocontents{#1}{#2}%
    \fi
}

\let\old@l@section\l@section
\renewcommand{\l@section}[2]{\old@l@section{\textbf{#1}}{\textbf{#2}}}
\makeatother

\appendix
\section*{%
  \texorpdfstring{%
    \raisebox{-.16\height}{%
      \includegraphics[width=0.06\textwidth]{figure/logo.png}%
    }%
    \hspace{0.5em}%
    \textbf{\gradientRGB{SpanVLA}{9,29,83}{23,99,180}}%
    \hspace{0.25em}Appendix%
  }{SpanVLA Appendix}%
}

{
    \hypersetup{linkcolor=black}
    \begin{spacing}{1.4} 
        \setcounter{tocdepth}{2} 
        \renewcommand{\contentsname}{} 
        
        \makeatletter
        \renewcommand\tableofcontents{%
            \@starttoc{stc}%
        }
        \makeatother
        
        \tableofcontents
    \end{spacing}
}

\vspace{0.7cm}

\section{Additional Related Work}
\paragraph{Reinforcement Fine-Tuning.}
Reinforcement fine-tuning (RFT) has improved general reasoning \citep{guo2025deepseek} and driving-policy optimization \citep{fu2025minddrive,rawal2026nord,jiang2025irl,zou2025diffusiondrivev2}. Most driving approaches build on expert-demonstration datasets without explicit real-world negative-recovery supervision. Early takeover-based methods, including TakeAD and CoReVLA, use preference optimization to learn from expert corrections collected in simulation \citep{liu2025takead,fang2025corevla,rafailov2023direct}. TakeVLA further explores reinforcement fine-tuning in reconstructed takeover scenarios \citep{gao2026learning}. These studies motivate learning from mistakes, while leaving real-world negative and recovery data largely unexploited. \spanvla addresses this gap through NR-RFT, which combines sample-type-specific penalties and rewards with an advantage correction for difficult rollout groups to stabilize the training. Together with nuReasoning-NR, it provides real-world data, a tailored training method, and a dedicated benchmark for studying negative-recovery learning.

\section{nuReasoning-NR Dataset and Benchmark}
\label{app:nureasoning_details}

nuReasoning-NR extends nuReasoning \citep{huang2026nureasoning} with real-world negative and expert recovery samples, providing the first real-world driving dataset and benchmark for learning from both forms of supervision. It contains 36K samples: 30K positive reasoning samples, 3K negative samples, and 3K expert recovery samples.
The three subsets cover common categories of interaction-intensive driving scenarios and provide complementary supervision for learning expert driving decisions, suppressing undesirable behaviors, and acquiring corrective actions.
They share a common data representation with eight surrounding camera streams, reference ego trajectories, object annotations, and HD maps, supporting both policy learning and planning evaluation.

\begin{figure}[t]
    \centering
    \includegraphics[width=\linewidth]{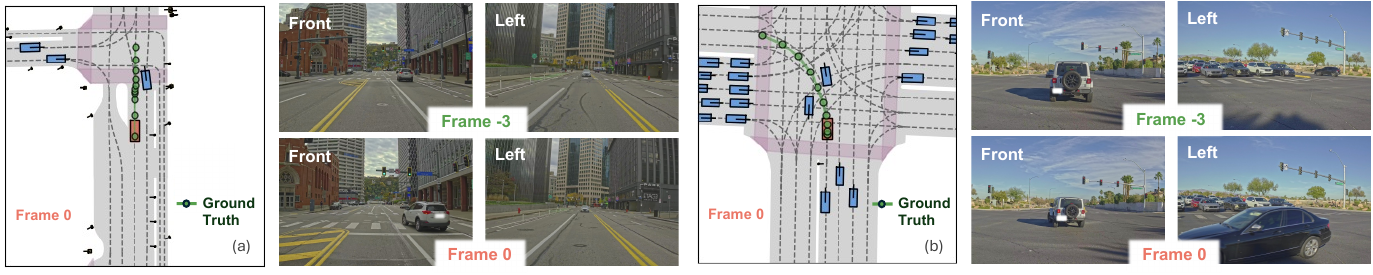}
    \caption{\footnotesize Complex scenario samples (positive samples) of nuReasoning-NR dataset. (a) Cut-in scenario: A vehicle from the front-right abruptly cuts in at the intersection. (b) Intersection scenario: The ego vehicle has already entered the intersection; despite the red traffic signal, it should yield appropriately and follow the leading vehicle to complete the left turn. Frame~$-3$ and Frame~0 denote the first history frame and the current frame for the camera streams.}
    \label{fig:interaction_scenarios}
\end{figure}

These samples provide trajectory-level supervision with distinct behavioral semantics.
Negative trajectories identify inefficient, hesitant, or potentially unsafe behaviors to suppress, while expert recovery trajectories specify corrective actions from difficult driving states.
Their event-level correspondence links suboptimal behavior to the subsequent expert correction, with each segment retaining its own timestamp and observation history.
This structure motivates the asymmetric supervision in NR-RFT.

\subsection{Scenario Coverage and Data Representation}

\paragraph{Scenario Coverage.}  We focus on complex and long-tail driving scenarios and collect 36K driving samples from Las Vegas, Boston, Pittsburgh, and Singapore, covering ego lane changes, lane bias, vulnerable road users (VRU), construction zones, stop signs, cut-in, and lead vehicle braking scenarios, as shown in \Cref{fig:data_distribution}. The history horizon of nuReasoning-NR dataset is 1.5 seconds, and the future horizon is 5 seconds (2 Hz) to include the key driving elements and events. Two typical complex driving scenarios of nuReasoning-NR dataset are shown in \Cref{fig:interaction_scenarios}.

\paragraph{Sensor Settings.} The vision data of nuReasoning-NR consists of eight surrounding camera streams, including front, front-left, front-right, left, right, back, back-left, and back-right. In addition, we provide annotations (e.g., trajectory, size, and category) for each object, e.g., vehicles, pedestrians, and traffic cones. Moreover, the HD map data is provided for these driving scenarios following the nuPlan map format \citep{karnchanachari2024towards} with driving lane, driving direction, and traffic light status.

\subsection{Real-World Negative and Recovery Data}

\paragraph{Event-Based Construction.} We curate negative and recovery samples from early-stage exploratory real-world driving tests. For each selected event, a negative segment records suboptimal ego behavior, and its corresponding recovery segment records the subsequent expert correction. Each segment forms a separate sample with its own observation history and reference trajectory. The resulting subsets contain 3K negative samples and 3K expert recovery samples. The correspondence is temporal and event-based: the negative and recovery segments occur at different moments and therefore have different observation inputs. They do not constitute two alternative trajectories conditioned on an identical driving state.

\paragraph{Supervision Semantics.}
Negative trajectories identify behaviors to avoid, including inefficient or hesitant maneuvers and potentially unsafe actions. Recovery trajectories demonstrate how an expert corrects behavior from a difficult driving state. Accordingly, a recorded negative trajectory serves as an undesirable reference, whereas an expert recovery trajectory serves as a desirable reference. These distinct meanings motivate the negative-behavior penalty and recovery-behavior reward in NR-RFT. Additional examples are shown in \Cref{fig:supp_negative_vis,fig:supp_recovery_vis}.

\subsection{Positive Reasoning Annotations}

\begin{figure}[t]
    \centering
    \includegraphics[width=\linewidth]{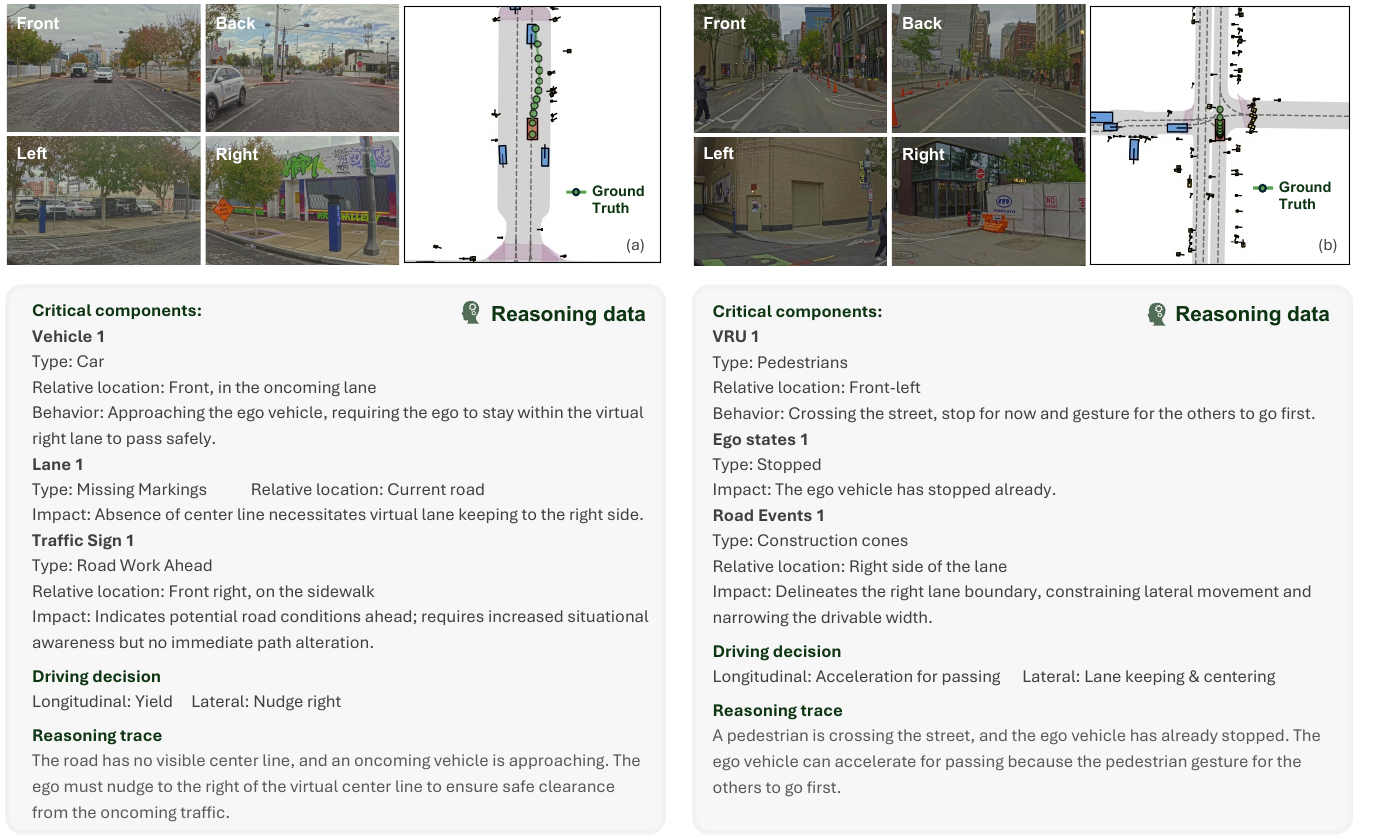}
    \caption{\footnotesize Reasoning data scenario samples (positive samples) of nuReasoning-NR dataset. (a) An oncoming vehicle encroaches into the current ego-vehicle's lane; the ego vehicle should nudge to the right to yield the oncoming vehicle. (b) A pedestrian gestures for the ego vehicle to proceed; since the ego vehicle has already come to a stop, it can safely start and pass.}
    \label{fig:CoT_scenarios}
\end{figure}

The positive subset provides expert trajectories and reasoning annotations for supervised fine-tuning. We use Gemini-3-Pro \citep{gemini3report2025} to generate structured annotations linking driving decisions to decision-relevant scene elements.
Representative scenarios and annotations are shown in \Cref{fig:interaction_scenarios,fig:CoT_scenarios}.

\paragraph{Annotation Inputs.}
The annotation model receives multi-view camera observations, ego states, and high-level driving instructions. Following \citet{zhou2025autovla}, recorded expert actions are also supplied as reference cues to encourage consistency between the explanations and demonstrated behavior. These future actions are used for offline annotation;
they are not provided as inputs to the driving policy at inference.

\paragraph{System Prompt and Output Structure.} The system prompt specifies the annotation VLM's role, annotation task, expected CoT reasoning format, and expert examples of CoT reasoning. The VLM role is an advanced full self-driving system, and the task is to determine the optimal driving decision based on the driving video from camera sensors and vehicle status inputs with CoT reasoning. Following the reasoning format of Alpamayo \citep{wang2025alpamayo}, we also leverage the compacted reasoning trace to keep the reasoning short and reduce redundant reasoning. Although the VLA model utilizes the short reasoning trace as the reasoning supervision signal, we still need a comprehensive reasoning output with critical components and driving decisions to ensure the reasoning quality. Moreover, we carefully provide some typical expert annotation examples for the annotation VLM as a reference. Then, the model requires generation with JSON format to reason critical components, driving decision, and reasoning trace one by one.

\paragraph{Quality Control.} We evaluate the reasoning quality based on the accuracy and completeness for the three parts: critical components, driving decision, and reasoning trace. The evaluation follows a strict binary protocol, and any error in each component results in a score of 0 for the corresponding part. Our human expert assessed 250 randomly selected samples, yielding an overall accuracy of 80.4\% in our complex scenarios. Erroneous samples were subsequently corrected or removed to construct the final high-quality reasoning dataset. \Cref{fig:CoT_scenarios} shows two typical CoT data in nuReasoning-NR dataset. 

\subsection{Benchmark and Evaluation Protocol}

\paragraph{Evaluation Subsets.}
The benchmark contains 300 negative samples and 300 recovery samples, all included in the 36K dataset total. Negative and recovery samples correspond to different timestamps, each with its own observation history and reference trajectory. During dataset curation, we prioritized scenario-level separation between training and evaluation to minimize overlap and avoid assigning negative and recovery segments from the same scenario to different splits. The negative subset evaluates planning in situations associated
with recorded suboptimal behavior. The recovery subset evaluates corrective planning from the difficult states in which expert recovery begins. Results are reported separately for the two subsets and averaged across the benchmark.

\paragraph{Evaluation Pipeline.}
We adapt the NAVSIM evaluation framework \citep{dauner2024navsim} to nuReasoning-NR using its object annotations and HD maps. The object trajectories, sizes, and categories, together with HD maps, enable candidate ego trajectories to be evaluated against surrounding traffic and road geometry.
This supports quantitative assessment of how negative and recovery supervision affects safety, progress, and comfort in challenging driving scenarios. The benchmark reports PDMS and its component metrics, covering collision avoidance, drivable-area compliance, ego progress, time-to-collision, and comfort. The reported benchmark scores measure planning performance; the additional negative and recovery reward terms are used for training. This evaluation supports systematic comparison of positive-only learning and learning with negative and recovery supervision.


\section{Details of SFT}
\subsection{Implementation Details}

SpanVLA supports mixed vision and textual prompt inputs. First, it takes multi-frame and multi-view image data as the vision inputs $\mathcal{V}^t = \bigcup_{i} \{ c_i^\tau \}_{\tau=t-T_h}^t$ where $c_i$ is each camera stream, which consists of four history frames per camera stream sampled at 2 Hz. For language inputs $\mathcal{T}^t=\{I^t, \{S_{\text{ego}}^\tau\}^t_{\tau=t-T_h}\}$, the system prompt describes the role and reasoning and planning task, and the high-level instruction $I$ (e.g., turn right), and the ego states $S_\text{ego}$ are inputted and tokenized as text. The latter consists of ego acceleration, history velocity, and trajectory.

Our SpanVLA model is trained with supervision finetuning via a two-stage paradigm. First, we leverage the CoT reasoning data and action tokens from the ground-truth trajectory to train the VLM backbone, enabling the backbone with the reasoning capability and aligning it with planning. We directly utilize and freeze the vision encoder of Qwen2.5-VL-3B and only train the LLM module. Then, the action expert is finetuned with the frozen VLM backbone without action token generation.

\subsection{Action Codebook} We leverage dual-representation for future action, where one is the continuous waypoint for the flow matching, and one is the discrete action token for autoregressive decoding. For the latter, we leverage the same codebook as \citep{zhou2025autovla}, comprising 2048 action tokens for 0.5 second vehicle motion from the Waymo Open Motion dataset \citep{ettinger2021large}. 

\section{Details of RFT with Negative-Recovery Samples}

\subsection{GRPO Objective}
\label{app:grpo_objective}
We introduce Negative-Recovery Reinforcement Fine-tuning (NR-RFT) to suppress negative driving behaviors and reinforce expert recovery actions. During training, we update the VLM backbone using positive, negative, and recovery samples. We adopt group relative policy optimization (GRPO)~\citep{shao2024deepseekmath,jiang2025alphadrive} for post-training.

Given a scenario query $q=(\mathcal{V}^t,\mathcal{T}^t)$, we sample $G$ candidate token sequences $\mathcal{Y}=\{y_1,\ldots,y_G\}$ from the old policy $\pi_{\theta_{\mathrm{old}}}$. Each $y_i$ contains reasoning when present and action tokens. Following the notation of the main paper, we denote the trajectory decoded from its action tokens by $o_i=\{\mathbf{p}_{i,t}\}_{t=0}^{T}$, where $\mathbf{p}_{i,t}\in\mathbb{R}^2$ is the ego center position at time step $t$. Policy probabilities and reasoning penalties are evaluated on $y_i$, while driving and reference-matching rewards are evaluated on $o_i$. We optimize the current policy $\pi_\theta$ using
\begin{equation}
\label{grpo_spanvla}
\begin{aligned}
\mathcal{J}_{\mathrm{GRPO}}(\theta)
&=\mathbb{E}_{q,\mathcal{Y}}
\Bigg[
\frac{1}{G}\sum_{i=1}^{G}
\Big(
\mathcal{J}_i^{R}
-\beta\mathbb{D}_{\mathrm{KL}}
\big(
\pi_\theta(\cdot\mid q)
\,\|\,
\pi_{\mathrm{ref}}(\cdot\mid q)
\big)
\Big)
\Bigg].
\end{aligned}
\end{equation}
The expectation is over training queries and output groups sampled from the old policy. The clipped reward objective is
\begin{equation}
\label{eq:grpo_terms}
\begin{aligned}
\mathcal{J}_i^{R}
=\min\Bigg(&
\frac{\pi_\theta(y_i\mid q)}{\pi_{\theta_{\mathrm{old}}}(y_i\mid q)}A_i,
&\operatorname{clip}\!\left(
\frac{\pi_\theta(y_i\mid q)}{\pi_{\theta_{\mathrm{old}}}(y_i\mid q)},
1-\epsilon,1+\epsilon
\right)A_i
\Bigg).
\end{aligned}
\end{equation}
Here, $\epsilon$ controls clipping, $\beta$ weights KL regularization, and $\pi_{\mathrm{ref}}$ is the reference policy obtained from SFT. Standard GRPO computes the advantage as
\begin{equation}
\label{eq:standard_grpo_advantage}
A_i^{\mathrm{GRPO}}
=\frac{r_i-\mu}{\sigma+\varepsilon_{\mathrm{num}}},
\qquad
\mu=\frac{1}{G}\sum_{j=1}^{G}r_j,
\end{equation}
where $r_i$ is the total reward associated with $y_i$ and its decoded trajectory $o_i$, $\sigma=\operatorname{std}(\{r_j\}_{j=1}^{G})$, and $\varepsilon_{\mathrm{num}}>0$ ensures numerical stability. NR-RFT uses the sample-type-dependent rewards defined below and replaces $A_i^{\mathrm{GRPO}}$ with the corrected advantage $A_i$ defined in the main paper. For the advantage correction, the all-zero condition is evaluated on the trajectory driving rewards $\{r_{\text{Driving},i}\}_{i=1}^{G}$ (PDMS or EPDMS), before applying the negative, recovery, and CoT terms.

\subsection{Reward Function}
Following the main paper, the total reward combines driving quality, negative-behavior suppression, expert recovery, and reasoning cost:
\begin{equation}
\label{eq:supp_nr_reward}
r_i
=r_{\text{Driving},i}
-\lambda_N r_{\text{Negative},i}
+\lambda_R r_{\text{Recover},i}
-\lambda_C r_{\text{CoT},i}.
\end{equation}
The non-negative coefficients $\lambda_N$, $\lambda_R$, and $\lambda_C$ weight the negative-behavior penalty, recovery-behavior reward, and CoT penalty, respectively. The trajectory-based terms are computed from $o_i$, and $r_{\text{CoT},i}=r_{\text{CoT}}(y_i)$ is computed from its associated reasoning. We define each term below.

\subsubsection{Driving Reward}

The reward function provides the primary training signal during RFT and guides policy optimization toward high-quality driving behaviors. To improve overall driving performance beyond merely imitating expert trajectories, we adopt a hybrid reward design in which the main driving reward, denoted by $r_{\text{Driving}}$, measures trajectory quality from the perspectives of safety, rule compliance, comfort, and progress.

For the NAVSIM v1 benchmark, we use the Predictive Driver Model Score (PDMS)~\citep{dauner2024navsim} as the primary driving reward. For the NAVSIM v2 benchmark, we use its extended version, the Extended Predictive Driver Model Score (EPDMS), which introduces additional rule-compliance metrics.
For the NAVSIM v1 setting, the driving reward is defined as

\begin{equation}
r_{\text{Driving}} = \mathrm{PDMS}, \quad \mathrm{PDMS}
= \mathrm{NC} \times \mathrm{DAC} \times
\left(
\frac{5\,\mathrm{TTC} + 2\,\mathrm{C} + 5\,\mathrm{EP}}{12}
\right),
\label{PDMS}
\end{equation}
where $\mathrm{NC}$ denotes No at-fault Collision, $\mathrm{DAC}$ denotes Drivable Area Compliance, $\mathrm{EP}$ denotes Ego Progress, $\mathrm{C}$ denotes Comfort, and $\mathrm{TTC}$ denotes Time-to-Collision. Following the standard definition, these sub-scores are normalized to $[0,1]$, and their weighted combination yields a comprehensive measure of driving quality. If the planner fails to produce a valid output, we assign a reward of $0$.

For the NAVSIM v2 benchmark, we adopt the Extended Predictive Driver Model Score (EPDMS) as the driving reward \citep{cao2025pseudo}. Compared with PDMS, EPDMS further incorporates Driving Direction Compliance (DDC) and Traffic Light Compliance (TLC), and replaces the single comfort term with more fine-grained comfort metrics, including Lane Keeping (LK), History Comfort (HC), and Extended Comfort (EC). Specifically, EPDMS is defined as
\begin{equation}
\begin{aligned}
r_{\text{Driving}} = \mathrm{EPDMS}
&=
\left(
\prod_{m \in \{\mathrm{NC},\mathrm{DAC},\mathrm{DDC},\mathrm{TLC}\}}
\mathrm{filter}_m(\text{agent}, \text{human})
\right) \\
&\quad \cdot
\left(
\frac{
\sum_{m \in \{\mathrm{TTC},\mathrm{EP},\mathrm{HC},\mathrm{LK},\mathrm{EC}\}}
w_m \, \mathrm{filter}_m(\text{agent}, \text{human})
}{
\sum_{m \in \{\mathrm{TTC},\mathrm{EP},\mathrm{HC},\mathrm{LK},\mathrm{EC}\}}
w_m
}
\right),
\label{EPDMS}
\end{aligned}
\end{equation}
Their corresponding weights are $w_{\mathrm{TTC}}=5$, $w_{\mathrm{EP}}=5$, $w_{\mathrm{HC}}=2$, $w_{\mathrm{LK}}=2$, and $w_{\mathrm{EC}}=2$.
Here, the filtering function is defined as
\begin{equation}
\mathrm{filter}_m(\text{agent}, \text{human})=
\begin{cases}
1.0, & \text{if } m(\text{human}) = 0,\\
m(\text{agent}), & \text{otherwise}.
\end{cases}
\end{equation}

For negative samples, we omit the ego-progress (EP) term from the driving reward because the recorded trajectory is an undesirable reference.

\subsubsection{Negative-Behavior Penalty and Recovery-Behavior Reward}

For each predicted trajectory $o_i$, we compute the negative-behavior penalty or recovery-behavior reward according to the sample type. Let $o^{\mathrm{ref}}=\{\mathbf{p}^{\mathrm{ref}}_t\}_{t=0}^{T}$ denote the reference trajectory associated with the current training sample. We define the average distance between $o_i$ and $o^{\mathrm{ref}}$ as
\begin{equation}
d(o_i, o^{\mathrm{ref}})
=
\frac{1}{T+1}\sum_{t=0}^{T}
\left\|
\mathbf{p}_{i,t}-\mathbf{p}^{\mathrm{ref}}_t
\right\|_2.
\end{equation}
Based on this distance, we define a unified reference-matching term
\begin{equation}
r_{\text{Ref}}(q,o_i)
=
\operatorname{clip}\left(
1-\frac{d(o_i,o^{\mathrm{ref}})}{\delta},
\,0,\,1
\right),
\end{equation}
where $\delta$ is a distance threshold. This term reaches its maximum value of $1$ when $o_i$ exactly matches the reference trajectory, decreases linearly as the deviation grows, and becomes $0$ when $d(o_i,o^{\mathrm{ref}})\ge \delta$.

For negative samples, $o^{\mathrm{ref}}$ corresponds to an undesirable behavior trajectory, and $r_{\text{Negative},i}=r_{\text{Ref}}(q,o_i)$ enters Eq.~\eqref{eq:supp_nr_reward} with coefficient $-\lambda_N$. For recovery samples, $o^{\mathrm{ref}}$ corresponds to a desirable recovery trajectory, and $r_{\text{Recover},i}=r_{\text{Ref}}(q,o_i)$ enters with coefficient $+\lambda_R$. Each term is zero for other sample types. The two terms share the same reference-matching function but use opposite signs to suppress undesirable behavior and reinforce expert corrections.

\subsubsection{Reasoning Penalty and Action-Reasoning Alignment}
To discourage unnecessarily long reasoning and achieve adaptive reasoning, we apply a length-based penalty on the generated reasoning. Let $L(y_i)$ denote the length in tokens of the CoT part in output $y_i$.
We define the CoT penalty as a sigmoid function:
\begin{equation}
\label{eq:cot_len_penalty}
r_{\text{CoT}}(y_i)=\frac{1}{1+\exp\!\left(-(L(y_i)-L_{\text{tol}})\gamma\right)},
\end{equation}
where $L_{\text{tol}}$ is the reasoning length at which the penalty equals $0.5$, and $\gamma>0$ controls the steepness. The penalty increases smoothly with reasoning length, discouraging unnecessarily long reasoning.

In addition to length regularization, we enforce an action-reasoning alignment reward.
Intuitively, the reasoning should be consistent with the high-level maneuver implied by the
predicted trajectory. We implement a rule-based detector that maps (i) the reasoning text in $y_i$ to
a maneuver label via keyword matching, and (ii) the predicted trajectory $o_i$ to a maneuver
label via simple geometric heuristics. We then penalize inconsistent pairs.

We extract a maneuver label $\hat{k}_{\text{text}}\in\mathcal{K}$ from the CoT via keyword matching, where
$\mathcal{K}$ is a small set of maneuver types (\textit{straight}, \textit{left turn},
\textit{right turn}, \textit{change lane to left}, \textit{change lane to right}, \textit{accelerate},
\textit{decelerate}). For instance, keywords such as ``straight'' or ``go forward'' map to
\textit{straight}; ``left turn'' or ``turn left'' map to \textit{left turn}; and ``accelerate'' or
``speed up'' map to \textit{accelerate}.

A scene may contain several actions and objects, although we assign it a single maneuver label. When several maneuver cues are found, we use a fixed priority order to select the label that best describes the main focus of the scene. The same priority order is used for both the reasoning and the trajectory. 
In parallel, we infer a maneuver label $\hat{k}_{\text{traj}}$ from the predicted trajectory $o_i$ using simple
geometric heuristics, including the signed heading change over the horizon for left/right/straight and
the speed change for accelerate/decelerate. Concretely, we compute the net heading change and net speed
change over the prediction horizon, and map them to maneuver types using fixed thresholds.
We then check alignment between $\hat{k}_{\text{text}}$ and $\hat{k}_{\text{traj}}$. If they are inconsistent, we apply a fixed penalty by overriding the CoT term, discouraging misaligned reasoning--action
pairs during RFT.

\subsection{Kullback--Leibler (KL) Divergence}
Following~\citep{zhou2025autovla}, we include a KL regularization term in the GRPO objective in Eq.~\eqref{grpo_spanvla} to constrain the updated policy $\pi_\theta$ to stay close to the
SFT reference policy $\pi_{\text{ref}}$. For each scenario query
$q=(\mathcal{V}^t,\mathcal{T}^t)$ and a sampled token sequence $y_i\in\mathcal{Y}$, the KL divergence is formulated as:
\begin{equation}
\mathbb{D}_{\text{KL}}(\pi_\theta \| \pi_{\text{ref}}) =
\frac{\pi_{\text{ref}}(y_i|q)}{\pi_\theta(y_i|q)} -
\log\!\left(\frac{\pi_{\text{ref}}(y_i|q)}{\pi_\theta(y_i|q)}\right) - 1,
\end{equation}
where $\theta$ denotes the parameters of the current policy $\pi_\theta$, and $\pi_{\text{ref}}$ is the
reference policy obtained from the SFT stage. This regularization penalizes large deviations from
$\pi_{\text{ref}}$, stabilizing policy updates and retaining useful knowledge learned during SFT while
optimizing for the driving reward.

\section{Additional Implementation Details}

The default configuration includes three camera views, i.e., front, front-left, and front-right, and can be easily extended to additional camera streams.


\section{Additional Experimental Results}
Our main SpanVLA RFT comparisons use a two-stage schedule with a total of three epochs. In the first epoch, we train exclusively on positive samples to establish a basic policy for successful driving. In the second and third epochs, we vary the training-data composition across the RFT variants: the positive-only variant continues using positive samples, the positive--negative variant uses both positive and negative samples, and the full NR variant uses positive, negative, and recovery samples. The three-epoch schedule is shared across these comparisons, with the data composition varied only in the second stage.
We report PDMS for NAVSIM v1 and EPDMS for NAVSIM v2, as defined in \Cref{PDMS,EPDMS}, respectively. The nuReasoning-NR diagnostic benchmark is evaluated using PDMS.

\subsection{Detailed Main Results}

\paragraph{NAVSIM Benchmarks.}
We report the detailed main results of SpanVLA. For a fair comparison, we only compare SpanVLA with methods that use neither additional information nor additional test-time scaling methods. \Cref{tab:navsim_v1} reports the results on the \texttt{navtest} benchmark of NAVSIM v1 against the state-of-the-art (SOTA) end-to-end driving models and VLA-based models. SpanVLA demonstrates strong performance in PDMS, and the RFT can significantly further improve the performance under the alignment with the reward signal. Moreover, as \Cref{tab:navsim_v2} illustrates, SpanVLA outperforms the SOTA methods on the \texttt{navtest} benchmark of NAVSIM v2. \Cref{tab:navhard} reports the SOTA VLA driving performance on the \texttt{navhard} benchmark.

\paragraph{nuReasoning Benchmark.}
We further evaluate SpanVLA on nuReasoning benchmark, which focuses on long-tail driving scenarios. The evaluation includes no-at-fault collision (NC), measuring the avoidance of collisions attributable to the ego vehicle; driveable-area compliance (DA), checking whether the ego footprint remains within the driveable area; ego progress (EP), measuring progress along the intended route relative to the ground-truth trajectory; comfort (CF), assessing compliance with acceleration, jerk, yaw-rate, and yaw-acceleration constraints; and human-likeness (HL), measuring similarity to human driving based on final displacement error. We also report average displacement error (ADE) over a 5-second horizon in meters. Higher values indicate better performance for all metrics except ADE. As shown in \Cref{tab:nuReasoning}, SpanVLA achieves the highest NPS of 73.3 and the lowest ADE of 1.4\,m among the compared methods.


\paragraph{nuReasoning-NR Benchmark.}
To evaluate planning specifically in negative and recovery scenarios, nuReasoning-NR provides two diagnostic test sets, each containing 300 scenarios. Among the methods in this comparison, only SpanVLA and AutoVLA are additionally trained with NR samples; all other compared models use positive data only. As shown in Table~\ref{tab:negative_recovery}, NR training improves both models over their positive-only counterparts on both test sets. The NR-trained variants also outperform all positive-only baselines on both sets, with SpanVLA achieving the highest PDMS on each subset and the highest average PDMS among the compared methods.

\begin{table}[t]
  \caption{\footnotesize Comparison with SOTA methods on the \textbf{NAVSIM v1} (\texttt{navtest}). PDMS (Predictive Driver Model Score), NC (No Collision), DAC (Drivable Area Compliance), EP (Ego Progress), TTC (Time-to-Collision), Comf. (Comfort). * denotes best-of-N selection with N = 6, where the candidate with the highest PDMS is chosen.}
  \vspace{0.2cm}
  \renewcommand{\arraystretch}{1.3}
  \label{tab:navsim_v1}
  \centering
  \scriptsize
  \setlength{\tabcolsep}{8.5pt}
  \begin{tabular}{l|cc|c|ccccc}
    \toprule
    Methods & Cam. & Lid. & PDMS $\uparrow$  & NC $\uparrow$ & DAC $\uparrow$   & EP $\uparrow$ & TTC $\uparrow$    & Comf. $\uparrow$ \\
    \midrule
    \rowcolor[HTML]{eaf1f8} \multicolumn{9}{l}{\textit{Conventional End-to-end-based Methods}} \\
    TransFuser \citep{chitta2022transfuser}      & \checkmark    & \checkmark & 84.0  & 97.8      & 92.6      & 78.9     & 92.9   & \textbf{100.0} \\
    DRAMA \citep{yuan2024drama}                  & \checkmark    & \checkmark & 86.9  & 98.2      & 95.2      & 81.3     & 94.2   & \textbf{100.0} \\
    Hydra-MDP \citep{li2024hydra}                & \checkmark    & \checkmark & 86.5  & 98.3      & 96.0      & 78.7     & 94.6   & \textbf{100.0} \\
    DiffusionDrive \citep{liao2024diffusiondrive} & \checkmark    & \checkmark & 88.1  & 98.2      & 96.2      & 82.2     & 94.7   & \textbf{100.0} \\
    WoTE \citep{li2025end}                        & \checkmark    & \checkmark & 88.3  & 98.5      & 96.8      & 81.9     & 94.4   & 99.9 \\
    \midrule
    \rowcolor[HTML]{eaf1f8} \multicolumn{9}{l}{\textit{VLA-based Methods}} \\
    ReCogDrive \citep{li2025recogdrive}            & \checkmark    & -    & 89.6  & 98.2      & 97.8      & 83.5     & 95.2   & 99.8 \\
    DriveVLA-W0 \citep{li2025drivevla}           & \checkmark    & -    & 90.2  & 98.7      & 99.1 & 83.3     & 95.3   & 99.3 \\
    AutoVLA \citep{zhou2025autovla}               & \checkmark    & -    & 89.1  & 98.4      & 95.6      & 81.9     & 98.0   & 99.9 \\
    \midrule
    \rowcolor[HTML]{fff3e9} \multicolumn{9}{l}{\textit{\textbf{Ours}}} \\
    SpanVLA (One-shot)                          & \checkmark    & -    &  82.1     &   97.5        &  90.8         &  76.9        &  93.7      &   99.5   \\
    SpanVLA (Post-RFT)                          & \checkmark    & -    &   90.8    &   99.6        &     99.1      &   81.1      &  98.1     &   \textbf{100.0}   \\
    
    SpanVLA*                          & \checkmark    & -    &   \textbf{93.3}    &   \textbf{99.8}        &     \textbf{99.8}      &   \textbf{85.0}       &  \textbf{99.2}     &   \textbf{100.0}   \\
    \bottomrule
  \end{tabular}
\end{table}

\vspace{-0.1cm}

\begin{table*}[t]
    \centering
    \setlength{\tabcolsep}{3.5pt}
    \caption{\footnotesize Comparison with SOTA methods on the \textbf{NAVSIM v2} (\texttt{navtest}). EPDMS (Extended Predictive Driver Model Score), NC (No Collision), DAC (Drivable Area Compliance), DDC (Driving Direction Compliance), TLC (Traffic Light Compliance), EP (Ego Progress), TTC (Time-to-Collision), LK (Lane Keeping), HC (History Comfort), EC (Extended Comfort). * denotes best-of-N selection with N = 6, where the candidate with the highest EPDMS is chosen.}
    \vspace{0.2cm}
    \label{tab:navsim_v2}
    \scriptsize
    \setlength{\tabcolsep}{6pt}
    \renewcommand{\arraystretch}{1.4}
    \begin{tabular}{l | c | c c c c c c c c c}
        \toprule
        Methods & EPDMS$\uparrow$ & NC$\uparrow$ & DAC$\uparrow$ & DDC$\uparrow$ & TLC$\uparrow$ & EP$\uparrow$ & TTC$\uparrow$ & LK$\uparrow$ & HC$\uparrow$ & EC$\uparrow$ \\
        \midrule
        \rowcolor[HTML]{eaf1f8} \multicolumn{11}{l}{\textit{Conventional End-to-end-based Methods}} \\
        TransFuser \citep{chitta2022transfuser} & 76.7 & 96.9 & 89.9 & 97.8 & 99.7 & 87.1 & 95.4 & 92.7 & \textbf{98.3} & 87.2 \\
        DiffusionDrive \citep{liao2024diffusiondrive} & 84.5 & 98.2 & 95.9 & 99.4 & \textbf{99.8} & 87.5 & 97.3 & \textbf{96.8} & \textbf{98.3} & \textbf{87.7} \\
        Hydra-MDP++ \citep{li2024hydra} & 81.4 & 97.2 & 97.5 & 99.4 & 99.6 & 83.1 & 96.5 & 94.4 & 98.2 & 70.9 \\
        Drivesuprim \citep{yao2025drivesuprim} & 83.1 & 97.5 & 96.5 & 99.4 & 99.6 & \textbf{88.4} & 96.6 & 95.5 & \textbf{98.3} & 77.0 \\
        ARTEMIS \citep{feng2025artemis} & 83.1 & 98.3 & 95.1 & 98.6 & \textbf{99.8} & 81.5 & 97.4 & 96.5 & \textbf{98.3} & - \\
        \midrule
        \rowcolor[HTML]{eaf1f8} \multicolumn{11}{l}{\textit{VLA-based Methods}} \\
        DriveVLA-W0 \citep{li2025drivevla} & 86.1 & \textbf{98.5} & 99.1 & 98.0 & 99.7 & 86.4 & 98.1 & 93.2 & 97.9 & 58.9 \\
        \midrule
        \rowcolor[HTML]{fff3e9} \multicolumn{11}{l}{\textbf{\textit{Ours}}} \\
        SpanVLA (One-shot) & 79.4 & 96.4 & 89.4 & 97.8 & \textbf{99.8} & 87.4 & 95.8 & 94.5 & 98.2 & 81.6 \\
        SpanVLA (Post-RFT) & 86.9 & 97.4 & 98.0 & 99.5 & \textbf{99.8} & 84.3  & 97.4 & 93.4 & 98.1 & 78.0 \\
        SpanVLA* & \textbf{90.5} & 98.4 & \textbf{99.4} & \textbf{99.7} & \textbf{99.8} & 86.9  & \textbf{98.3} & \textbf{96.8} & 98.2 & 78.9 \\
        \bottomrule
    \end{tabular}
\end{table*}

\begin{table*}[t]
    \centering
    \setlength{\tabcolsep}{5pt}
    \caption{\footnotesize Comparison with SOTA methods on the \textbf{NAVSIM v2} (\texttt{navhard}). * denotes best-of-N selection with N = 6, where the candidate with the highest EPDMS is chosen.}
    \vspace{0.2cm}
    \label{tab:navhard}
    \scriptsize
    \setlength{\tabcolsep}{4pt}
    \renewcommand{\arraystretch}{1.1}
    \begin{tabular}{l | c | c | c c c c c c c c c}
        \toprule
        Methods & Stage & EPDMS$\uparrow$ & NC$\uparrow$ & DAC$\uparrow$ & DDC$\uparrow$ & TLC$\uparrow$ & EP$\uparrow$ & TTC$\uparrow$ & LK$\uparrow$ & HC$\uparrow$ & EC$\uparrow$ \\
        \midrule
        \rowcolor[HTML]{eaf1f8} \multicolumn{12}{l}{\textit{Conventional End-to-end-based Methods}} \\
        \multirow{2}{*}{LTF \citep{chitta2022transfuser}} & 1 & 23.1 & 96.2 & 79.6 & 99.1 & 99.6 & 84.1 & 95.1 & 94.2 & \textbf{97.6} & \textbf{79.1} \\
         & 2 & 23.1 & 77.8 & 70.2 & 84.3 & 98.1 & 85.1 & 75.7 & 45.4 & 95.8 & \textbf{76.0} \\
        \midrule
        \multirow{2}{*}{RAP \citep{feng2025rap}} & 1 & 36.9 & 97.1 & 94.4 & 98.8 & 99.8 & 83.9 & 96.9 & 94.7 & 96.4 & 66.2 \\
         & 2 & 36.9 & 83.2 & 83.9 & 87.4 & 98.0 & \textbf{86.9} & 80.4 & 52.3 & 95.2 & 52.4 \\
        \midrule
        \rowcolor[HTML]{fff3e9} \multicolumn{12}{l}{\textbf{\textit{Ours}}} \\
        \multirow{2}{*}{SpanVLA (Post-RFT, Positive Only)}& 1 & 40.1 & 98.4 & 94.3 & 97.8 & 99.9 & \textbf{85.7} & 97.2 & 94.2 & \textbf{97.6} & 72.1 \\
        & 2 & 40.1 & 86.9 & 84.3 & 87.1 & 98.2 & 85.5 & 82.7 & \textbf{62.3} & \textbf{96.8} & 67.4 \\
         \midrule
        
        \multirow{2}{*}{SpanVLA (Post-RFT with NR)} & 1 & 40.8 & 99.1 & 94.2 & 99.8 & \textbf{100.0} & 66.6 & 98.4 & 91.8 & 97.3 & 47.1\\
         & 2 & 40.8 & 89.8 & 84.5 & \textbf{94.8} & 98.6 & 63.4 & 88.2 & 53.1 & 95.2 & 50.8\\
         \midrule
        \multirow{2}{*}{SpanVLA*} & 1 & \textbf{47.9} & \textbf{99.8} & \textbf{99.1} & \textbf{99.9} & \textbf{100.0} & 71.4 & \textbf{99.3} & \textbf{97.8} & 97.3 & 49.8\\
         & 2 & \textbf{47.9} & \textbf{90.9} & \textbf{86.8} & 93.8 & \textbf{99.1} & 69.5 & \textbf{89.4} & 54.5 & 95.6 & 50.9\\
        \bottomrule
    \end{tabular}
\end{table*}

\subsection{Additional Ablation Studies for Action Bridging}
\begin{wrapfigure}{R}{0.28\linewidth}
    \centering
    \vspace{-1.3cm}
    \includegraphics[width=0.28\textwidth]{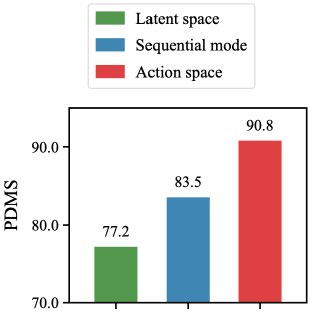}
    \caption{\footnotesize Comparison of three bridge modes for SpanVLA.}
    \vspace{-1cm}
    \label{fig:bridge_mode}
\end{wrapfigure}

Action bridging with flow matching serves a dual role: (i) extracting information from the VLM backbone via the KV cache, and (ii) predicting the flow-matching vector field. Accordingly, we instantiate three variants of Action Bridging with Flow Matching, differing in how bridging is formulated and where the vector field is predicted. (1) Action-space (SpanVLA). The bridging network directly predicts the vector field in the action space. We encode the history trajectory with an MLP to produce ten action-space queries, which are then used for vector-field estimation. (2) Latent-space. We use an MLP to encode the history and future trajectories into a latent feature, and perform vector-field prediction entirely in this latent space. (3) Sequential mode: We first use the bridging module to produce a conditional feature, and then run multi-step flow matching in the action space conditioned on this fixed feature. As \Cref{fig:bridge_mode} illustrates, the action mode, which fuses and predicts the vector field within the action space, outperforms the latent-space and sequential variants.

\begin{figure}[b]
    \centering
    \includegraphics[width=0.95\linewidth]{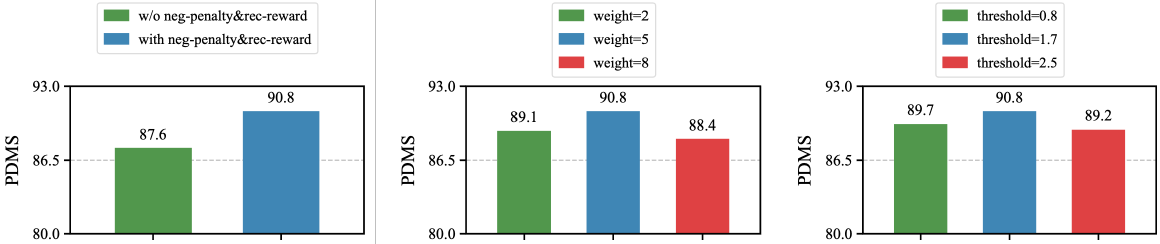}
    \caption{\footnotesize Ablation studies on negative-behavior penalty and recovery-behavior reward. 
        \textbf{Left}: With vs.\ without the negative/recovery reward terms.
        \textbf{Middle}: Varying the weight of the negative/recovery terms.
        \textbf{Right}: Varying the proximity threshold of the negative/recovery terms.}
    \label{fig:l2_ablations}
\end{figure}

\subsection{Additional Ablation Studies for RFT}

\begin{figure}[t]
    \centering
    \vspace{0.3cm}
    \includegraphics[width=\linewidth]{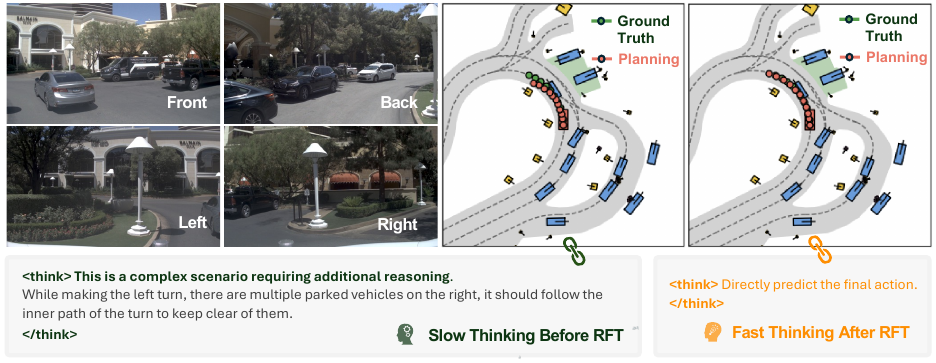}
    \caption{\footnotesize Additional comparison of planning and reasoning performance of SpanVLA model in \textbf{positive samples} of nuReasoning-NR dataset before and after RFT.}
    \label{fig:supp_positive_vis}
\end{figure}

\begin{figure}[t]
    \centering
    \vspace{0.3cm}
    \includegraphics[width=\linewidth]{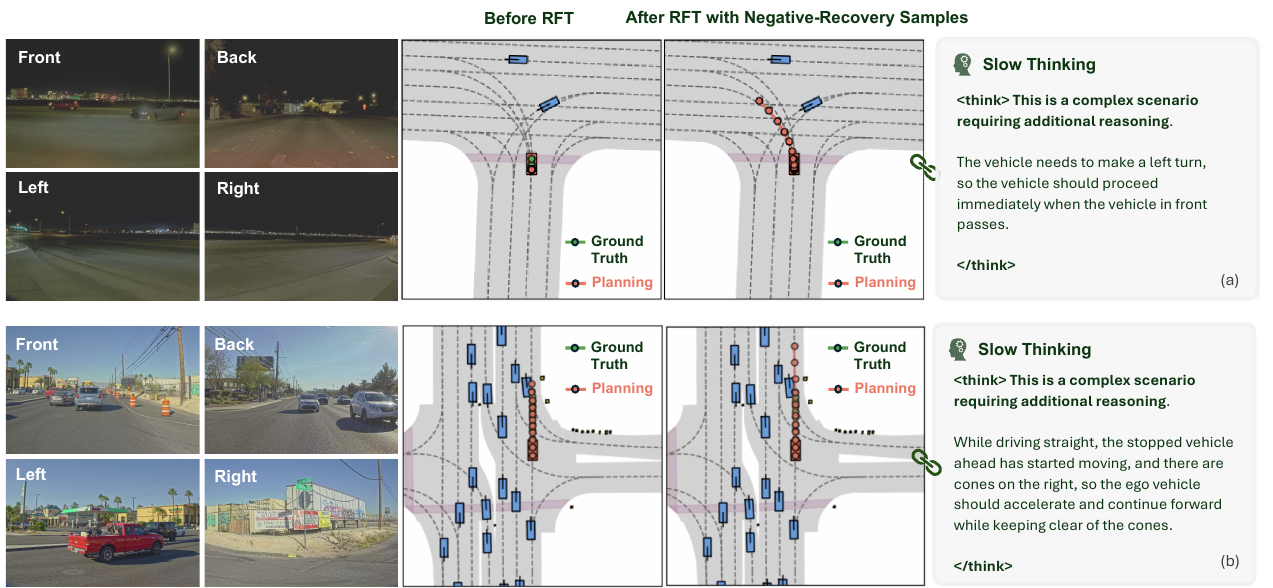}
    \caption{\footnotesize Additional comparison of planning and reasoning performance of SpanVLA model in \textbf{negative samples} of nuReasoning-NR dataset before and after RFT with negative-recovery samples.}
    \label{fig:supp_negative_vis}
\end{figure}

\subsubsection{Effect of Negative Penalty and Recovery Reward}
\label{app:neg-penalty-rec-reward}
We evaluate the negative-behavior penalty and recovery-behavior reward along three axes: the presence of these terms, their weights, and the proximity threshold that controls their activation. These experiments use positive, negative, and recovery training samples. \Cref{fig:l2_ablations} summarizes the results.

\paragraph{Presence of the Reward Terms.}
We compare training with a PDMS-only reward against training with the negative-behavior penalty and recovery-behavior reward. Removing the two terms corresponds to setting $\lambda_N=\lambda_R=0$. Including them improves PDMS from 87.6 to 90.8, showing a benefit from the distance-based reward terms beyond the driving reward in this setting.

\paragraph{Reward-Weight Sensitivity.}
The negative-behavior penalty and recovery-behavior reward are weighted by $\lambda_N$ and $\lambda_R$, respectively, following Eq.~\eqref{eq:supp_nr_reward}. The middle panel of \Cref{fig:l2_ablations} compares different weight settings. The moderate setting achieves the highest PDMS among the tested configurations, while both smaller and larger settings yield lower scores.

\paragraph{Proximity-Threshold Sensitivity.}
The threshold $\delta$ controls the range over which the negative-behavior penalty and recovery-behavior reward are active: the reference-matching term is nonzero when the average L2 distance to the corresponding reference trajectory is smaller than $\delta$. It also controls how the term decays with distance within this range. The right panel of \Cref{fig:l2_ablations} compares $\delta\in\{0.8,1.7,2.5\}$, which yield PDMS values of 89.7, 90.8, and 89.2, respectively. The intermediate threshold performs best among the tested values.

\subsection{Additional Qualitative Results}
\label{app:qualitative_results}
\begin{figure}[t]
    \centering
    \vspace{0.3cm}
    \includegraphics[width=\linewidth]{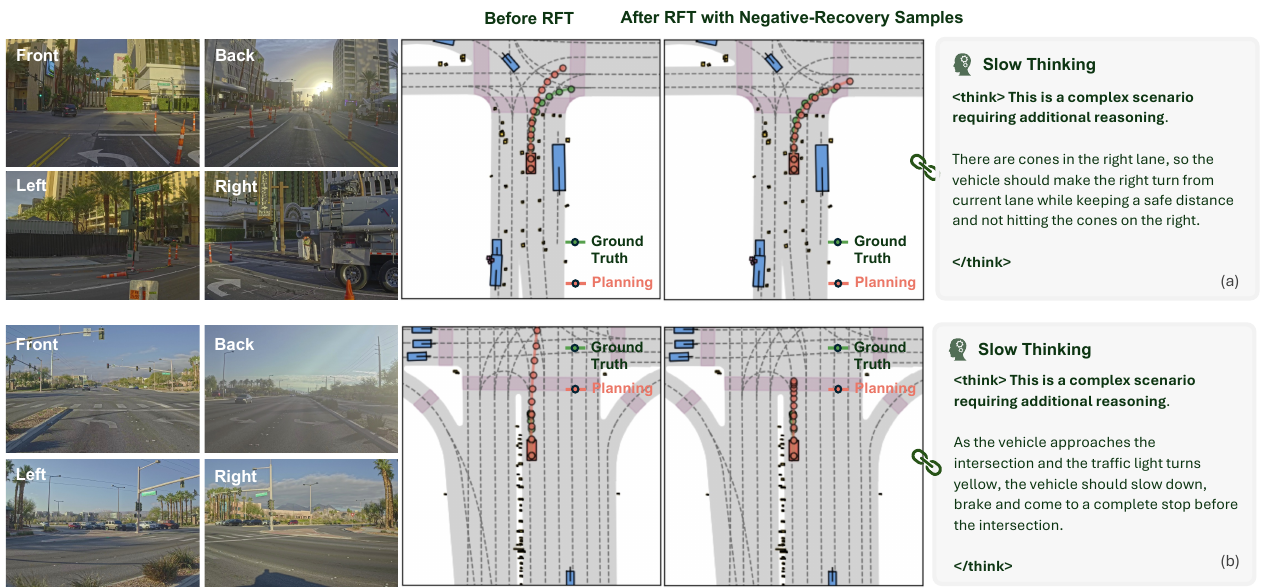}
    \caption{\footnotesize Additional comparison of planning and reasoning performance of SpanVLA model in \textbf{recovery samples} of nuReasoning-NR dataset before and after RFT with negative-recovery samples.}
    \label{fig:supp_recovery_vis}
\end{figure}

\Cref{fig:supp_positive_vis} shows the results with positive samples, and the RFT can reduce the unnecessary reasoning in simple scenarios like \citep{zhou2025autovla}.
\Cref{fig:supp_negative_vis} and \Cref{fig:supp_recovery_vis} present additional qualitative results for negative and recovery samples. Negative samples help the vehicle avoid overly conservative, yield-induced stopping behaviors during turning (\Cref{fig:supp_negative_vis} (a)) or going straight (\Cref{fig:supp_negative_vis} (b)), enabling it to proceed with left turns or accelerate to continue straight. Recovery samples further demonstrate how the vehicle can execute a right turn by temporarily using the only available left-turn lane when the right-turn lane is closed due to a construction zone, as shown in \Cref{fig:supp_recovery_vis} (a), and still brake to a stop when a yellow light transitions to red, as shown in \Cref{fig:supp_recovery_vis} (b). 

\section{Limitations and Future Work}
Although our efficient action bridging significantly accelerates inference, the current runtime of 1.5\,Hz remains insufficient for direct real-world deployment. Notably, we have not yet applied hardware-level optimization or deployment acceleration. For reference, Alpamayo \citep{wang2025alpamayo} achieves 1.75\,ms per token under optimized deployment, whereas our current implementation requires 33\,ms per token. Such optimization strategies are orthogonal to our method and can be directly applied to our model. As future work, we plan to deploy the model on a real-world platform with system-level acceleration. In addition, while our experiments demonstrate that negative-recovery samples substantially improve policy performance and robustness, designing more effective reward functions to fully unlock the potential of these data remains an open problem and an important direction.

\section{Acknowledgements}
The authors would like to express their gratitude to Qian Zhu, Haram Kim, and Baoshu Qi for their extensive efforts in data preparation and annotation of the nuReasoning-NR dataset. Special thanks also go to Muhammad Taufik Tirtosudiro and Jiong Yang for their support in developing the evaluation pipeline. The authors also thank Nitin Kapania, Sourabh Vora, and Balajee Kannan for their strong support for the project. Moreover, the authors want to thank the Amazon AI PhD Fellowship and the Qualcomm Innovation Fellowship for their support.

\end{document}